\documentclass[journal,twoside]{IEEEtran}
\usepackage{graphicx}
\usepackage{amsmath}
\usepackage{amssymb}
\usepackage{algorithm}
\usepackage{algorithmic}
\usepackage{booktabs}
\usepackage{subfigure}
\usepackage{stfloats}
\usepackage{nccmath}
\usepackage{bm}
\usepackage{multirow}
\usepackage{array}
\usepackage{url}

\begin{document}

\newcommand{\tabincell}[2]{\begin{tabular}{@{}#1@{}}#2\end{tabular}}
\newcommand{\PreserveBackslash}[1]{\let\temp=\\#1\let\\=\temp}
\newcolumntype{C}[1]{>{\PreserveBackslash\centering}p{#1}}


%

\title{Face Aging Effect Simulation using Hidden Factor Analysis Joint Sparse Representation}

\author{Hongyu~Yang,~\IEEEmembership{Student Member,~IEEE,}
        Di~Huang,~\IEEEmembership{Member,~IEEE,}
        Yunhong Wang,~\IEEEmembership{Member,~IEEE,}    
        Heng~Wang,            
        and~Yuanyan Tang,~\IEEEmembership{Fellow,~IEEE}

\thanks{H. Yang, D. Huang, Y. Wang, and H. Wang are with the Laboratory of Intelligent Recognition and Image Processing, Beijing Key Laboratory of Digital Media, School of Computer Science and Engineering, Beihang University, Beijing 100191, China (e-mail: hongyuyang@buaa.edu.cn, dhuang@buaa.edu.cn, yhwang@buaa.edu.cn, hengwang@cse.buaa.edu.cn).}
\thanks{Y. Tang is with the Department of Computer and Information Science, Faculty of Science and Technology, University of Macau, Taipa 853, Macau (e-mail: yytang@umac.mo).}
}



\maketitle

\begin{abstract}
Face aging simulation has received rising investigations nowadays, whereas it still remains a challenge to generate convincing and natural age-progressed face images. In this paper, we present a novel approach to such an issue by using hidden factor analysis joint sparse representation. In contrast to the majority of tasks in the literature that handle the facial texture integrally, the proposed aging approach separately models the person-specific facial properties that tend to be stable in a relatively long period and the age-specific clues that change gradually over time. It then merely transforms the age component to a target age group via sparse reconstruction, yielding aging effects, which is finally combined with the identity component to achieve the aged face. Experiments are carried out on three aging databases, and the results achieved clearly demonstrate the effectiveness and robustness of the proposed method in rendering a face with aging effects. Additionally, a series of evaluations prove its validity with respect to identity preservation and aging effect generation.

\end{abstract}


\ifCLASSOPTIONpeerreview
\begin{center} \bfseries EDICS Category: 3-BBND \end{center}
\fi
%
\IEEEpeerreviewmaketitle

\section{Introduction}
\IEEEPARstart{H}{uman} faces convey rich information of the individual's identity, age, gender, ethnicity, emotion, \emph{etc}. During the last few decades, great efforts have been devoted to machine based face image analysis, mainly deriving from its wide potential applications in the real world and the rapid advances of computer vision and machine learning techniques. Exactly as the studies on the traditional issues, {\em i.e.} lighting or pose changes, the attempts on age variations also constitute a significant branch of this field, involving age invariant face recognition \cite{10,17,18}, aging simulation\footnote{We use aging simulation, aging synthesis, and age progression alternately in this paper.} \cite{3,6,9}, age estimation \cite{14,20,32}, \emph{etc}. Particularly, the task on face aging simulation has been given increasing attention in these years, since the solution to this complex issue benefits many attractive applications \cite{1,2,13}:

\subsubsection{Multi-Media and Entertainment}Along with the flourish of the film industry, generating visual effects by computer has become essential and dominant. The aging and rejuvenating processing on the actors' faces via computed-aided approaches makes it possible to achieve fantastic rendering to the audiences without consuming so much time and material resources. 

With regard to the ordinary people, face beautification softwares are more familiar, some of which are among the most popular ones in the mobile application store. For example, smoothing skin wrinkles and amending facial configurations closely depend on the face rejuvenation techniques. Likewise, appearance prediction of an old age would inspire people's curiosity and interest as well.

\subsubsection{Aging Compensation in Face Analysis} Aging is always considered as a challenge in face recognition, and aging effect simulation thus deserves thorough studies to compensate such variation in automatic face recognition systems. Furthermore, face aging synthesis could also enormously benefit forensic art, face image retrieval across age, automatic update of face databases, and provide useful references for seeking the missing individuals.

\subsection{Related Work}
Due to the attraction by the aforementioned real-world applications, the attempts on face aging simulation have experienced a gradual transition from computer graphics to computer vision. According to the studies in \cite{14,16, 29},  human face age progression can be generally summarized as two stages, {\em i.e.}, child growth and adult aging: the skeletal growth plays a dominant role from infancy to grown-up, while the texture details (\emph{e.g.} wrinkles) distinguish seniors from young adults. Inspired by these observations, some approaches based on crania development theory and skin wrinkle analysis have been investigated in recent years, and the previous work mainly develops in three directions:

\subsubsection{Coordinate Transformation based}
The early efforts to address the issue of human face aging synthesis mainly focused on skin's anatomy structure and facial muscle changes, and some physical measurements and anthropometry driven methods were proposed. In \cite{16,36}, a computable growth model of human heads was thoroughly introduced by Todd  {\em et\ al.}, and it was simulated in a sequence of profiles generated by a variant of the revised cardioidal-strain transformation model. Based on the skin's anatomy structure, Wu {\em et\ al.} \cite{37,38} presented a 3-layered skin model, and the wrinkles were obtained by the relative motion between layers with the definitized interlaminar constraints satisfied. O'Toole {\em et\ al.} \cite{25} calculated the average of a certain number of sample faces captured by the 3D laser scanner, and by adjusting the distance between any test face and the average, the face appeared older or younger. Ramanathan {\em et\ al.} \cite{7} presented a craniofacial growth model for facial appearance prediction of young people, characterizing growth related shape variations by means of the parameters defined over a set of facial landmarks, and further extended their work for modeling adult age progression \cite{15}. The mechanical synthesis methods are dedicatedly designed; however, they are computationally complex, and there is a shortage of realness of the synthesized faces.  

\subsubsection{Texture Transplanting based}
With such approaches \cite{15, 35}, age progression is conducted by transferring the age-related details to the given test face. Shan {\em et\ al.} \cite{34} presented an image based method to transfer the geometric details from one surface to another. By this method, both face aging and rejuvenating simulation were achieved. Tiddeman {\em et\ al.} \cite{3} claimed that the aged texture from a single training image leads to a lack of statistical validity and the unrealism of the synthesized face, and they thus captured the mean differences between the images of two age groups. Kemelmacher {\em et\ al.} \cite{9} followed this idea and further took another factor into special consideration, {\em i.e.}, illumination. In order to match the lighting, they projected the input image into every age subspace by computing a rank-4 basis via singular value decomposition on the images of this age group. Apparently, it is a direct way to remove the unpleasant stochastic texture by averaging a number of faces; the age related  high-frequency details are smoothed out simultaneously, however. In \cite{5,6}, Suo {\em et\ al.} introduced a compositional and dynamic graph model for face aging. For the face images in each age group, it used a hierarchical And-Or graph for description, whose nodes denoted the decomposed parts of the face, and face aging was then modeled as a Markov process on the parse graph representation. Whereas the gallery and test faces utilized in this model were collected under strictly controlled environment, and its robustness to other variations in practice is thus problematic. The idea of texture transplanting based methods is straightforward and the implementation is generally not complex. Nevertheless, the identity information cannot be well preserved in the synthesized result due to the replacement of key facial regions.

\subsubsection{Aging Function based}
These approaches are mostly example-based, and involve in another significant and closely related issue, {\em i.e.}, age estimation. The aging function reveals the relationship between the facial texture and the age. Lanitis {\em et\ al.} \cite{4} built a statistical face model in which the distinctive aging patterns for different individuals were learnt, and they established unique simulation for the given probe face. In \cite{33}, long-term age progression was modeled by connecting sequential short-term patterns following the Markov property of the aging process, and the function-based method was exploited to extract these short-term appearance changes. Park {\em et\ al.} \cite{10,11} extended this kind of approaches to the 3D space, and applied a 3D-assisted face model to offset the texture and shape changes of probe faces caused by aging. With the view-invariant 3D face model transformed from a given 2D face image, they separately considered the shape and texture changes, and the aging pattern was approximated by a weighted average of the ones in the training set. The super-resolution techniques in a tensor space were firstly introduced to face aging synthesis by Wang {\em et\ al.} \cite{12,13}. Different from the previous work building regression functions between the face image and its age, they established a relationship between a down-sampled test face image and a high-resolution one in which the age-related features were added. The aging function based approaches usually possess good universality and suit for the most test faces; however, due to the diversity of the aging patterns and insufficient training data, the aging functions cannot be accurately defined, and the obtained aging effects are not evident or touching enough.

\subsection{Challenges}
In spite of the encouraging progress in face aging simulation, it still remains a challenge to synthesize ideal age transformations due to some issues as follows:

\subsubsection{Complexity in Aging} 
Human face aging involves in both the common rules and distinct patterns. People share similar age-related changes, and the aggravation of wrinkles and the growth of profile included; whereas the diverse genetic factors introduce the stochasticity and varieties for different individuals, making the aging processes differ from each other. Furthermore, a number of external factors, also have impact on it. For example, lifestyle would retard or accelerate the aging rate of one person, and the makeups together with accessories would lead to deviation between the facial appearance and its actual age. All of these causes raise the uncertainty during age progression and the difficulty in appearance prediction.

\subsubsection{Data Collection}
Most of the recent age progression approaches are data-driven, where the aging patterns and age-related features are learnt from the training samples. Data collection is thus extremely crucial to generate such statistical models. Only with the premise that enough aging variations and skin details are covered at the training stage, the synthesis results are significative and credible. Either due to lacking of images from the elderly people, or being pressed for long-term aging sequence from a single individual, the existing publicly available facial aging databases are far from sufficient.
  
\subsubsection{Other Interferences} 
Another reason that makes aging synthesis a challenging task is that the practical probe faces usually undergo variations in expression, pose, and lighting; hence a robust aging model is supposed to take them into account. Due to the existence of the unsolved complex aging model mentioned above, most of the other interferences are still strictly restrained in current aging simulation studies.

\subsection{Motivation and Contribution}

\begin{figure*}[!t]
\centering
\includegraphics[width=0.97\textwidth]{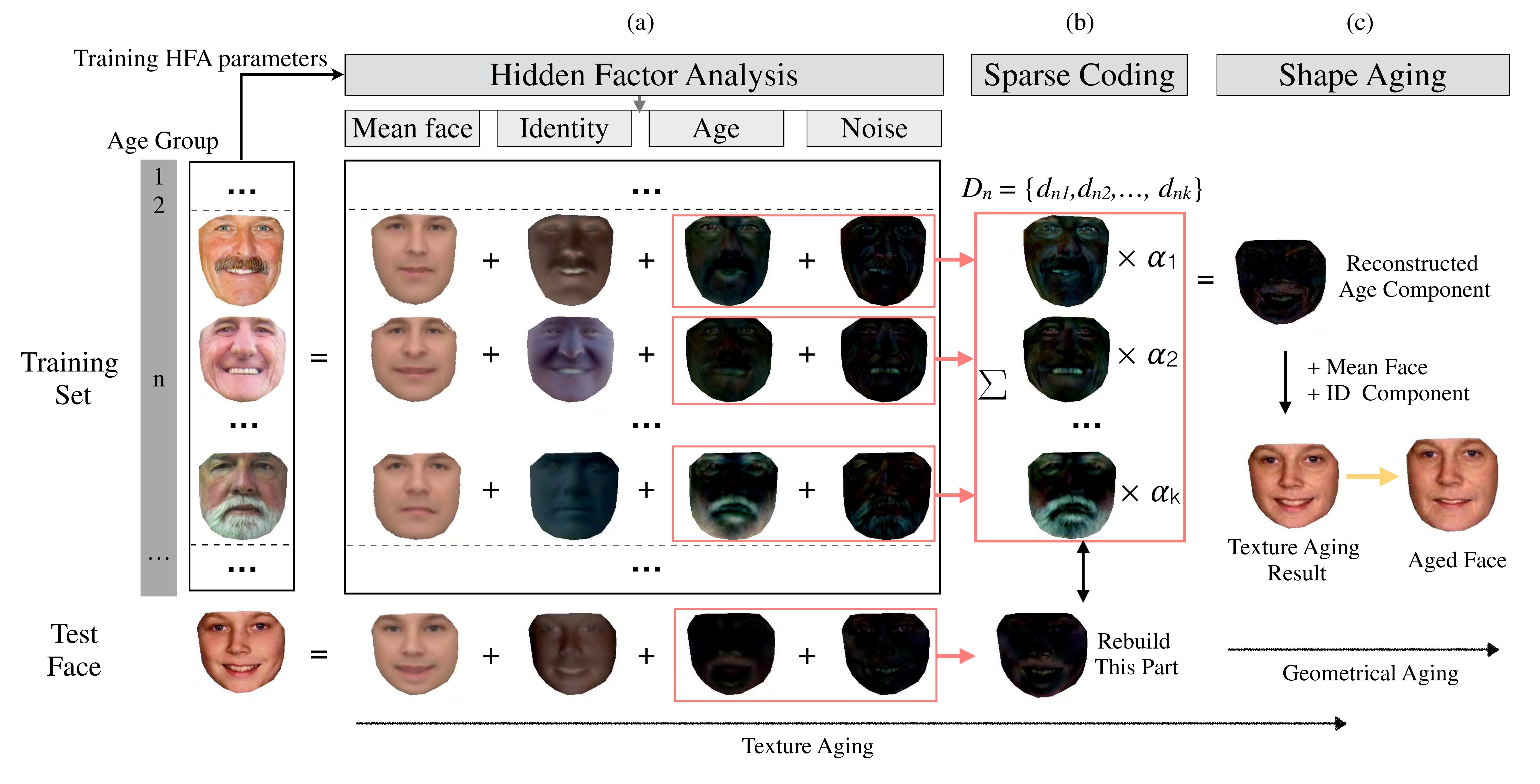}
\caption[justification=centering]{Illustration of the proposed aging approach. At the training stage, the training faces together with their labels are applied to learn the bases of the age and identity subspace. (a) Visualization of the diverse facial components achieved based on the well trained Hidden Factor Analysis (HFA) model; (b) The age components from the target age group constitute a dictionary for reconstructing that of the test face with the aging effects added; (c) The shape aging transformation is further conducted on the obtained texture aging results to synthesize the craniofacial growth.}
\label{framework}
\end{figure*}

Face age progression aims to render a given face image with aging or rejuvenating effects on it, and moreover, the identity information is supposed to be well preserved during synthesis. Indeed, human faces convey abundant information, such as gender, emotion, pose and identity, and the age attribute is one of them. For such a specific task of aging simulation, the multiple attributes presented by faces can be mainly summarized as two categories, {\em i.e.} the age-related and age-invariant attributes, according to if they are stable over time. The traditional aging approaches basically learn the aging pattern parameters explicitly, or produce the age-related characteristics by simply averaging the training samples belonging to a specific age group, without considering the disturbance of features that are intrinsically not relevant to age, {\em e.g.} identity. These facts and the aspiration for better aging rendering motivate us to find ways to accurately separate these mixed attributes presented on a face, {\em i.e.}, the person-specific component which mainly conveys the identity information, and the age-related component that is inextricably bound to age. Only the latter tends to change in the process of aging simulation.

When the multi-attribute decomposition is settled, we then need to transform the age-related part to a target age group. As introduced in Section {\em I.A}, the majority of the texture transplanting and aging function based approaches generally formulate the relationship between two age groups, and apply the fixed trained aging parameters or a templated wrinkle mask to any of the test images. Allowing for the individuality conveyed by each test face, however, this strategy is apparently not qualified. When there is a large range of variety within the test set, such as skin color, illumination, {\em etc.}, few work following this trend in the literature manages reconciling the aging effects of the synthesized face and its consistency with the input one, and customized aging transformation is thus needed. 

Based on the above analysis, we propose a novel approach to address the issue of face aging synthesis, by using Hidden Factor Analysis (HFA) joint sparse representation. The HFA is first exploited to extract the age-related and person-dependent properties from the assorted facial attributes. And aging transformation is then achieved on the obtained age component by sparse representation, supposing that the input facial age component can be linearly combined by a certain number of prototype elements that possess similar architecture, and append desired age-related semantic meaning to the new representation. Since a fitting term is usually predefined in the cost function during sparse coding, the information of the source face is expected to be preserved in the new one, indicating that the reconstruction better fits the given data itself. Finally, the reconstructed age component is combined with the original person-dependent component, to generate the age-progressed facial image. We test the aging method on three databases, {\em i.e.} FG-NET, MORPH, and IRIP, including both aging and rejuvenating synthesis, and satisfactory results are achieved, especially for handling the expression and skin color variations. Additionally, a series of evaluations prove its effectiveness regarding the ability of identity preservation and age synthesis.

The main contribution of this work is three-fold:
\begin{itemize}
\item To address the common problem of data shortage, we collect a database consisting of more than 2,000 face images from Internet with age ranging from 1 to 70, and 68 landmarks are manually labeled on each face.
\item We propose a novel and general approach to face aging effect simulation, separately modeling the person-specific properties that tend to be stable in a relatively long period and the cues that change gradually over time.
\item We adapt sparse representation to the issue of face aging effect synthesis, which better reconstructs the age related component by accounting for its consistency with the input face, making the synthesized texture more natural.
\end{itemize} 

\subsection{Organization of this Paper}
The rest of this paper is organized as follows: In Section II, the proposed aging approach framework is introduced. Section III provides a detailed description of our facial texture aging simulation method. The shape aging pattern is presented in Section IV. Section V displays and analyzes the experimental results on three databases, along with some discussion in Section VI, and Section VII concludes this paper.

\section{Approach Framework}
In this section, the framework of the proposed face aging simulation method is introduced.

Fig. 1 shows an illustration of the proposed aging approach. Given a set of face images from diverse age groups, for the purpose of settling the multi-attribute decomposition problem, a latent factor analysis method, {\em i.e.} HFA, is introduced into our method. All the training samples together with their labels are applied to learn the bases of the age and identity subspace. Once the parameters are well trained, by projecting the training and test faces onto these two subspaces, the age-related and personalized component can be achieved respectively for every face image. Additionally, all the faces also share a common part, {\em i.e.} the mean face, and a noise part. Theoretically, the age-related properties entirely lie in the age component, whereas the noise part actually contains some cues that correspond to the subtler wrinkles and texture details in higher frequency. To avoid some of these age-specific information being lost and keep the generated face consistent with the source image in terms of lighting condition, expression, \emph{etc.}, the age and noise part are seen as a whole in the following procedure. At the next stage, the sparse representation technique is adopted to transform the age element to a target age group. As shown in Fig. 1 (b), the obtained age components of the training samples of the target age group constitute a dictionary for sparse reconstruction, and by the linear combination of these atoms approximate to that of the test image, the coefficients are recovered. Thus the desired age component can be generated by a weighted sum of the atoms with corresponding age properties. Then substituting the reconstructed age component for the original one in the test face yields an aged rendering. The facial texture aging simulation is finally achieved, followed by the shape aging transformation. We calculate the mean profiles of every age cluster, and apply the difference between them onto the test face to synthesize the shape changes as age grows. Fig. 2 shows a brief demonstration of the age progression results.

\section{Texture Aging Effect Simulation}
Face age progression includes both the skeletal growth and texture senility. In this section, the texture aging synthesis approach is described in detail. Briefly speaking, it contains extracting the age-related facial component from the mixed attributes via HFA, and utilizing sparse representation to rebuild it to generate the aging effects. Additionally, patch-based face representation which refines the reconstructed results is introduced at the end of this section.

\begin{figure}[t]
\centering
\includegraphics[width=3.45in]{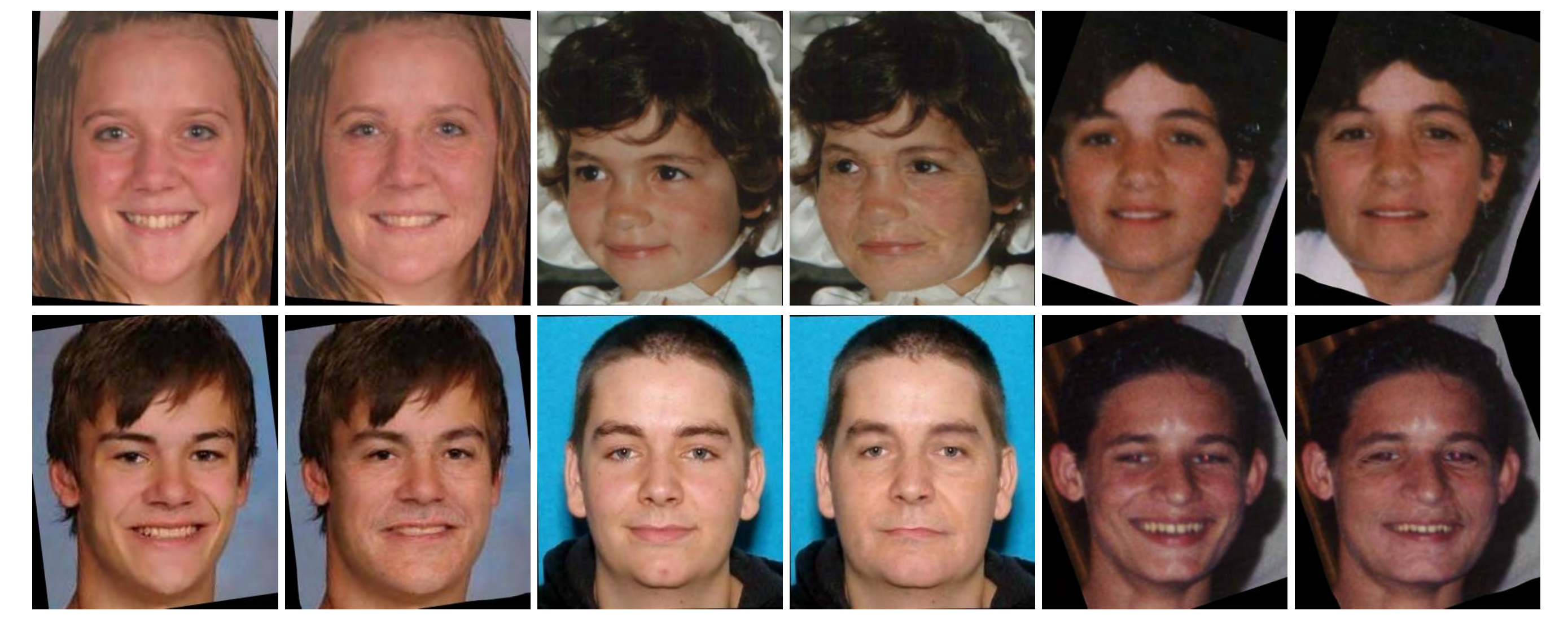}
\caption{Demonstration of the aging synthesis results.}
\label{demo}
\end{figure}
\subsection{Age-related Property Extraction based on HFA}
Human faces carry various information, and for the purpose of simplifying the representation in face aging synthesis, they can be generally summarized as the age-related and person-specific attributes. The diversification of these attributes makes it possible to distinguish faces from each other, and a face can be thus denoted as a dependent variable effected by age and identity. This assumption is applicable not only for age progression, but also for some other face analysis, such as automatic age estimation. In \cite{18}, Gong {\em et\ al.} assumed that a face can be expressed as a linear combination of the age component and identity component, and presented the HFA method to address the age invariant face recognition problem. HFA is introduced into our aging model to separate the person-specific facial properties and age-specific characteristics, which are assumed as statistically independent, and the latter one will be altered in our model for generating aging effects on faces. Specifically, a face is formulated as:
\begin{equation}
\large
\vec{f} = \vec{m} + U\vec{x} + V\vec{y} + \vec{\varepsilon}
\label{eq:projection}
\end{equation}

The major notations used are defined as follows:

\begin{itemize}
\item $\vec{f}$: \textit{d} $\times$ 1 dim, the observed face. 
\item $\vec{m}$: \textit{d} $\times$ 1 dim, the mean value over the whole training faces.
\item \textit{U}: \textit{d} $\times$ \textit{p} dim, whose columns span the identity subspace.
\item \textit{V}: \textit{d} $\times$ \textit{q} dim, whose columns span the age subspace.
\item $\vec{x}$: \textit{p} $\times$ 1 dim, the latent identity factor with prior distribution of ${N} (0, I)$.
\item $\vec{y}$: \textit{q} $\times$ 1 dim, the latent age factor with prior distribution of ${N} (0, I )$. 
\item $\vec{\varepsilon}$: \textit{d} $\times$ 1 dim, the additive noise, following an isotropic Gaussian where $\vec{\varepsilon}\sim$\textit{N} (0,~$\sigma^{2}I$).
\end{itemize}

The mutual configuration and texture properties that the whole faces possess are defined as the mean face in this formulation; contrarily, the identity component $U\vec{x}$, age component $V\vec{y}$, and the additive noise $\vec{\varepsilon}$ representing other variations endow individuality on each of the observed faces, and make them distinctive. According to (1), the mean face can be computed by:
\begin{equation}
\large
\vec{m} = \frac{1}{N}\sum\nolimits_{i,j} \vec{f}_{ i}^{ j}
\label{eq:projection}
\end{equation}
where $N$ is the number of the training set. The identity and age subspaces in this linear representation are initially unknown, whereas the corresponding labels of the training samples are definite, according to which the faces can be grouped. In detail, the ordered pair $\langle\vec{f}_{i}^{j}$, $\vec{x}_{i}$, $\vec{y}_{j}$$\rangle$ represents the observed face of the $i$-$th$ subject at $j$-$th$ age group, $\vec{x}_{i}$ and $\vec{y}_{j}$ are the identity and age factors respectively. In order to seek the hidden parameters $\theta = \{\vec{m}, U, V, \sigma^{2}\}$ in this model, an objective function based on the maximum likelihood estimation is defined, and the parameters can be learnt by maximizing it:
\begin{equation}
\large
Lc = \sum\nolimits_{i,j}~\mathrm{ln} ~{p_{\theta}}  ({\vec{f}_{i}^{j}}, ~{\vec{x}_{i}}, ~{\vec{y}_{j}})
\label{eq:projection}
\end{equation}

As we know, with the given model parameter $\theta$, the posterior distribution of the latent variables ${p_{\theta_{0}}}({\vec{x}_{i}}, {\vec{y}_{j}}|F)$ can be estimated; conversely $\theta$ can be updated by maximizing the expectation of $L$. It should be noted that $\vec{x}_{i}$ and $\vec{y}_{j}$ are unobservable latent variables here; thereby the coordinate ascent approach is adopted, for alternately updating the parameters and variables in the model while with the others fixed. More specifically, given an initial estimation of the parameters $\theta_{0}$, a new estimation can be obtained by maximizing:

\begin{equation}
\large
\langle Lc \rangle = \sum_{i,j} \int {p_{\theta_{0}}} ({\vec{x}_{i}}, ~{\vec{y}_{j}}|F)~\mathrm{ln} ~{p_{\theta}}  ({\vec{f}_{i}^{k}}, ~{\vec{x}_{i}}, ~{\vec{y}_{j}}) d{\vec{x}_{i}} d{\vec{y}_{j}}
\label{eq:projection}
\end{equation}

To optimize such an issue involving in hidden variables as (4), the Expectation Maximization (EM) algorithm is exploited here. Given the training set $F = \{\vec{f}_{i}^{j}| i = 1,...N; j = 1,...M \}$ and an initial parameter estimation $\theta_{0}$, the conditional moments of $p_{\theta_{0}}(\vec{x}_{i}|F)$, $p_{\theta_{0}}(\vec{y}_{j}|F)$ and $p_{\theta_{0}}(\vec{x}_{i},\vec{y}_{j}|F)$ can be calculated at the Expectation stage; then at the Maximization stage, it updates the parameters $\theta = \{\vec{m}, U, V, \sigma^{2}\}$ with the obtained sufficient statics $\langle\vec{x}_{i}\rangle$ and $\langle\vec{y}_{j}\rangle$, satisfying that the new estimator maximizes the expectation of the log-likelihood function defined in (4). 

The EM algorithm jointly estimates both the latent factors and subspaces from a set of training face images, and samples of the same individual or age group share identical latent factor $\langle\vec{x}_{i}\rangle$ or $\langle\vec{y}_{j}\rangle$. In order to derive a better optimization of parameters $\theta$, it maximizes the expectation of the defined objective function so that the given faces are generated from this model with the maximum possibility. After several iterations the algorithm converges. With the well trained parameters $U$ and $V$, the identity and age components of any given face $\bm{f}$ can be calculated then, we only present the conclusions here:

\begin{equation}
\large
U\vec{x} = UU^{T} \Sigma^{-1} (\vec{f}-\vec{m})
\label{eq:projection}
\end{equation}

\begin{equation}
\large
V\vec{y} = VV^{T} \Sigma^{-1} (\vec{f}-\vec{m})
\label{eq:projection}
\end{equation}
where $\Sigma$ = $\sigma^{2}$I +$UU^{T}$+$VV^{T}$.

HFA was originally proposed to address the classification problem, and the vector $\bm{f}$ represented the extracted features. In our model, $\bm{f}$ denotes the observed face image. Fig. 1 (a) visualizes some of the face decomposition results of the male subjects, and more examples of the female subjects are shown in Fig. 3. We can see that HFA separately models the person-specific and age-specific attributes of the given face image in individual components. The personalized characteristics, such as the appearance, facial feature profile, and skin color, all distribute in the identity component; while the properties of higher frequencies, such as the white beard (in Fig. 1) and wrinkles around eyes, mainly lie in the age component. The flexibility of attribute decomposition offers more accurate analysis in further processing.

\begin{figure}[t]
\centering
\includegraphics[width=3.4in]{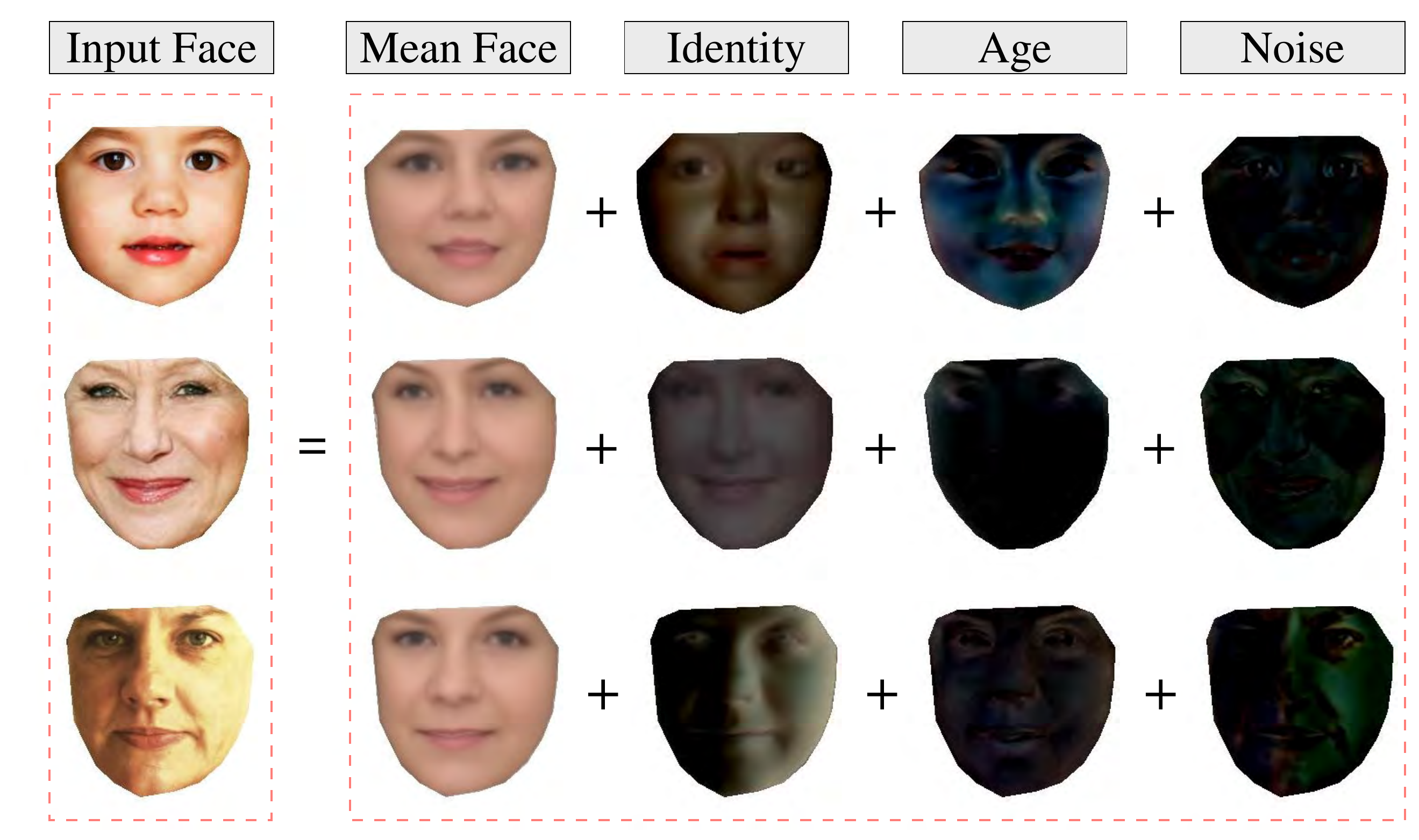}
\caption[justification=centering]{Examples of facial component decomposition via HFA. }
\label{HFA}
\end{figure}

It should be noted that to achieve face alignment and denote the whole faces as $d$-dimensional vectors, Active Appearance Model (AAMs) \cite{8} is adopted to warp them to a unified shape and size; and we can convert the obtained facial components back likewise. Specifically, before training the HFA model, all the face images are normalized to the average shape, and all the obtained decomposition components in (1) are in a unified shape as well.

The most explicit way is to warp the age component ($V\bm{y} + \bm{\varepsilon}$) of the training set and that of the test face from the mean shape to the original test shape for the following sparse reconstruction; however, we do not do this as the facial wrinkles will be repeatedly distorted, which would cause malformation of the synthesized wrinkles as well as lack of fidelity of the achieved aged faces. Instead we denote the age component ($V\bm{y} + \bm{\varepsilon}$)  as ($\bm{f} - \bm{m} - U\bm{x} $), and this formulation helps to eliminate the weaknesses of AAMs. In particular, the two parts of mean face $\bm{m}$ and identity $U\bm{x}$  are warped back to the original shape and further subtracted from the source face image $\bm{f}$, and the remaining part is exactly the age component. If ignoring the weaknesses of AAMs, the age component achieved in this way is the same as warping the one calculated by ($V\bm{y} + \bm{\varepsilon}$) . However, the former effectively helps to reduce the warping operation on high-frequency skin folds and wrinkles and thus avoid the malformation caused by repeated shape transformations. We then could deform the age components in the training set to the same shape of the test face and generate aged effects, and the age-related part only undergoes deformation for just one time through the whole process.

\subsection{Face Texture Aging Synthesis via Sparse Representation}
When the age component that changes gradually over time is separated out from the mixed facial attributes, we further transform it to a target age subspace, generating aging effects. The most direct and traditional measure is to average the photos of each age cluster and compute the difference between the input and target group. However, this method would degrade under complex conditions, and it cannot bridge the gap among diverse races and skin colors. Moreover, some of the age related details would encounter abrasion during averaging. Based on such consideration and the inspiration from the theory of sparse representation, that some of the signals, such as images, have naturally sparsity with respect to fixed bases and hence exhibit degenerate structure \cite{21}, in this paper we exploit sparse representation to address the issue of aged facial texture synthesis. Sparse representation techniques have been used as a powerful tool for many computer vision tasks in recent years, such as face recognition \cite{19,22}, image super-resolution \cite{26}, image and video restoration \cite{23, 27}, and signal compression \cite{28}. In our case, it shows the ability to render aging effects on a given probe face by representing its age component by a sparse linear combination of the gallery ones which convey the desired age properties.

The sparse representation based method mainly contains two parts in general, namely construction of gallery dictionary and computation of sparse coefficients. Traditionally, an over-complete dictionary is learnt from the given data based on a task-specific prior term in the loss function, whereas in our case, the stage of dictionary learning via minimizing the certain predefined loss function is superseded by HFA. After the multi-attribute decomposition step, we obtain $K_{i}$ age components for the $i$-$th$ age cluster, denoted as vectors $\bm d_{i_{j}}$ $( j = 1,2,...,K_{i})$ being of $d$-dimensional, and they constitute the dictionary:

\begin{equation}
\large
D_{i} = \{ \bm{d}_{i_{1}},\bm{d}_{i_{2}},...,\bm{d}_{i_{K_{i}}}\} \in \mathcal{R}^{ d \times K_{i} }
\label{eq:projection}
\end{equation}

Owing to the above-mentioned sparsity nature of signals, given the prepared dictionary $D_{i}$, any new age component from a young age group also denoted as a $d$-dimensional vector $\bm{y} \in \mathcal{R}^{ d \times 1}$, can be approximately represented as a linear combination of the elements in $D_{i}$. The basic reconstruction constraint is that the rebuilt age component is supposed to be concordant with the probe, and we thus formulate the aging process as:
\begin{equation}
\large
\widehat{\bm{\alpha}}_{0} = \mbox{arg min}\|\bm{\alpha}\|_{0} ~~\mbox{subject to}~~ \bm{y} = D_{i}\bm{\alpha}
\label{eq:projection}
\end{equation}
where $\|\cdot\|$ denotes the $\ell^0$ norm of vector $\bm{\alpha}$, counting the number of non-zero entries in it, and the obtained $D_{i}\widehat{\bm{\alpha}}_{0}$ is the expected age component. Since the dictionary $D_{i} $ merely contains age components from the $i$-$th$ age group, the corresponding age properties are spontaneously involved in the reconstructed result  $D_{i}\widehat{\bm{\alpha}}_{0}$, and selecting distinct dictionary $D_{i} $ makes the transformed face image appear younger or older, where the residual error exactly reflects the age-specific divergence between different age brackets. Many algorithms based on convex optimization or greedy pursuit are available to solve (8), and in this study, we adopt the homotopy algorithm to recover the linear relationship. 

Sparse representation helps leading in the desired characteristics with a designed dictionary, and keeping the reconstructed age component consistent with the original one. Whereas that's not the only benefit; because a term enforcing sparsity is involved during finding representations, the potential concern that the reconstruction error is probably close to zero and no aging effects are added can thus be avoided by controlling the sparsity of $\alpha$. Therefore substituting the reconstructed age component $D_{i}\widehat{\bm{\alpha}}_{0}$ for the original one of the test face yields aged rendering.

\subsection{Patch-based Representation for Face Modeling}

\begin{figure}[t]
\setcounter{figure}{3} 
\centering
\includegraphics[width=3in]{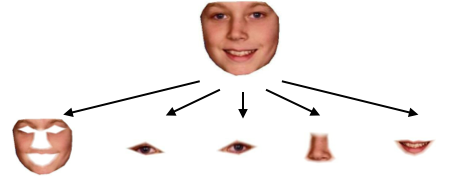}
\caption{Patch-based representation for face modeling.}
\label{PatchBasedRepresentation}
\end{figure}

Biological aging, muscle motion, and gravity effects would all lead to skin changes \cite{1}. These changes include some subtle wrinkles and creases as well as the integral skin sagging, such as dropping cheek, lower eyelid bags and deep grooves at the junction of facial features. Directly applying the above presented aging algorithm and handling the face as a whole would cause blurring in the zones of facial features, especially for the faces with exaggerated expressions. Due to such concerns, we further improve our aging algorithm with a patch-based scheme of face representation to pursue better aging synthesis results.

Then another thing worth discussing follows, {\em i.e.}, the granularity of dipartition. In comparison with the areas of facial features, aging effects are more concentrated in the zone of skin mask; moreover, the global downward trend would be lost if the particle size is excessively small. On the other hand, age progression for facial features cannot be ignored either, since minor texture changes effect much on the aging rendering as well. For example, the expressions in the eyes of seniors are prone to be cloudy, while those of children are much pure; hence merely adding wrinkles on cheeks would cause a lack of fidelity and consistency. Consequently, the patch-based representation in our work contains the subregions of eyes, nose, mouth and skin mask excluding them, as shown in Fig. 4, and age component reconstruction is individually conducted on these subregions.

\section{Shape Aging Pattern}
\begin{figure*}[!t]
\setcounter{figure}{4} 
\centering
\includegraphics[width=0.9\textwidth]{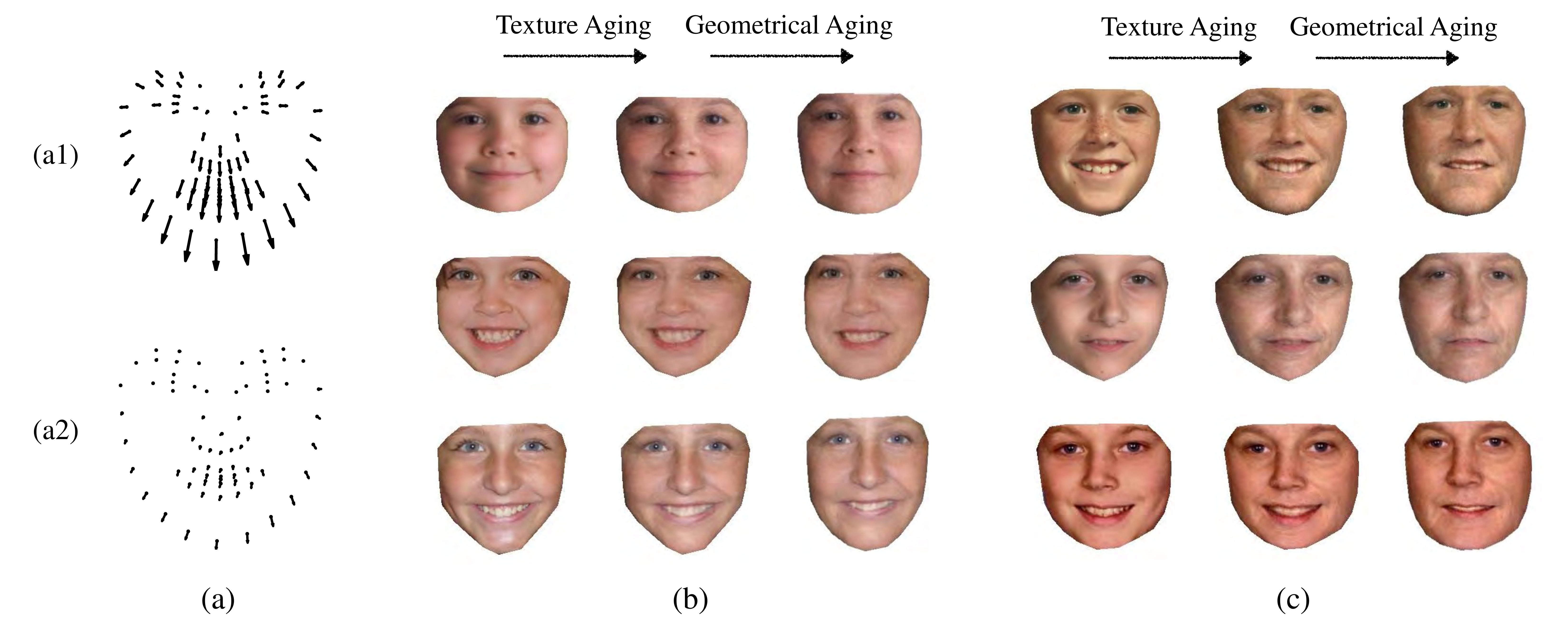}
\caption[justification=centering]{Illustration of the shape aging model. (a1) and (a2) present the geometrical deformation from child and young adult of male subjects to seniors respectively; (b) and (c) demonstrate some of the shape aging results. }
\label{shapeAging}
\end{figure*}

\begin{table*}[t]
\renewcommand{\arraystretch}{1.3}
\caption{Comparison of the aging databases in Gender, Ethnicity and Age Distrubution}
\label{Databases}
\centering
\begin{tabular}{c|c|c|c|c|c|c|c|c|c|c}
\hline
\hline
\multirow{2}{*}{\bfseries{Database}} & 
\multirow{2}{*}{\bfseries{Gender}} &
\multirow{2}{*}{\bfseries{Ethnicity}}& 
\multirow{2}{*}{\bfseries{Distribution}}&\multicolumn{7 }{c}{\bfseries{Age Group}}  \\
\cline{5-11} & &   &  & \bfseries{1} & \bfseries{2} & \bfseries{3}  & \bfseries{4} & \bfseries{5} & \bfseries{6} & \bfseries{7+} \\
\hline
\multirow{2}{*}{FG-NET} & \multirow{2}{*}{Male and Female}  &  \multirow{2}{*}{Caucasian}  &   \# Images  & 371 &  339 & 144 & 79 & 46  & 15  & 8\\
\cline{4-11} &  & & Distribution (\%) & 37.03 & 33.83   & 14.37 & 7.88 &  4.59 &  1.50 & 0.80 \\
\hline
\multirow{2}{*}{MORPH} & \multirow{2}{*}{Male and Female} & \multirow{2}{*}{\tabincell{c}{Caucasian, Hispanic, \\Asian, or African American}} &  \# Images  & 0 &  8205 &  14356 & 15217 &  11090 & 2973 &  258 \\
\cline{4-11} & & &Distribution (\%)  & 0 & 15.75 &27.56 & 29.21 & 21.29 & 5.71 & 0.50 \\
\hline
\multirow{2}{*}{IRIP} &  \multirow{2}{*}{Male and Female} & \multirow{2}{*}{\tabincell{c}{Caucasian, Asian, \\or African American}} &   \# Images & 300  & 300  & 300  & 300 & 300 & 300 & 300\\
\cline{4-11} & &  & Distribution (\%) & 14.29 & 14.29 & 14.29 & 14.29  & 14.29  & 14.29  & 14.29  \\
\hline
\hline
\end{tabular}
\end{table*}

When the texture aging effects are generated on faces, we further process the obtained intermediate aging results to synthesize the craniofacial growth.

In order to achieve shape aging effects, a statistical learning-based method is adopted as in \cite{5}. In detail, we calculate the mean profiles over each age cluster, and apply the difference between them ($S_{target} -S_{input}$) onto the test face to synthesize shape transformation as age grows:
\begin{equation}
\large
S_{new} = S_{subject} + (S_{target} -S_{input})
\label{shapeAging}
\end{equation}

Fig. 5 (a) visualizes this cranium growth model, in which the facial contours are denoted by the coordinates of a set of facial key-points, (a1) and (a2) demonstrate the geometrical deformation from child and young adult of male subjects to seniors respectively. The length and orientation of the arrows describe the magnitude and direction of skull growth. We can see that our statistical results exactly coincide with the `revised' cardioidal strain transformation described in \cite{7, 16}, that dramatical morphological changes arise as infancy grows to adulthood, while the facial configuration remains almost invariable during adult aging.

\section{Experimental Results}
To evaluate the effectiveness and generality of the proposed method of face aging simulation, we carry out extensive experiments on the IRIP database, which is collected by ourselves, and the publicly available FG-NET \cite{39} and MORPH  \cite{40} aging databases. The datasets, experimental protocol, and aging results are introduced in the subsequent.

\subsection{Database}
\subsubsection{Publicly available Datasets}
The FG-NET aging database is the most popular one for facial aging analysis, including aging simulation, age estimation, and age invariant face recognition. It contains 1,002 color or gray face images from 82 individuals with variations in expression, pose, and lighting condition (13 images per person on average). The age of faces varies from 0 to 69, 640 images are from the individuals less than 18 years old, and the maximum age gap within a person is 54 years. MORPH is a large database which includes two sections, {\em i.e.}, album 1 and album 2. It records subject's ethnicity, height, weight and gender. An extension of MORPH contains 52,099 images, the average age is 33.07, and the longitudinal age span of a subject ranges from 46 days to 33 years.

Tab. 1 apparently shows that only 69 face images in the FG-NET database are from the people older than 40 years old; even though the MORPH database contains more than 50,000 images, no image under 16 years old is included, and in the elderly age group 7 (more than 61 years old), only 17 images are from the female while 241 from the male, whereas most of which are African American, making its distribution uneven in age, gender and race. 

\footnotetext[2]{The IRIP database is available at \url{http://irip.buaa.edu.cn/new/data/irip_aging_database.zip}}
\subsubsection{IRIP Database\protect\footnotemark[2]}
In order to obtain sufficient training samples, especially for the seniors, we collect a face database named IRIP from the Internet with people from different age groups. This database contains 2,100 high-resolution color images in the wild whose size is about 400 $\times $ 500 in pixel. It covers a wide range of population in terms of age, race and gender. Ninety percent of the images are from the Caucasian, and the Asian and Africa American each accounts for about five percent. For every single age ranging from 1 to 70, 30 face images are collected, distributing evenly on male and female. One third of the male subjects are with beard, and around 50 faces in the database wear glasses. Additionally, 68 landmarks are manually labeled on every face for the purpose of pose alignment and shape aging modeling.

\subsection{Experimental Settings}

It is commonly acknowledged that human facial appearances share similar texture and configuration properties during a certain period. On the other hand, as the training faces within each age cluster grow in number, the sample variance increases as well, therefore we attempt to predict the aged facial look within an age interval, rather than that of a specific age. Particularly, we divide the face images into seven age groups, each of which covers 10 years, in our experiment: [1,10], [11,20], [21-30], [31-40], [41-50], [51-60], [61-70]. Because of lack of images of elderly people in the FG-NET and MORPH databases, we employ the IRIP database as the training set, and carry out face aging synthesis on all of them. For the sake of eliminating the interference caused by gender, we conduct appearance prediction apart for male and female; hence for every training age group, 150 face images are included.

Prior to training the age progression model, the normalization step is employed. All the faces are segmented from the background and aligned to an average shape of 100 $\times $ 100 in pixel by using AAMs. They are grouped into forementioned seven age groups, and each face sample is assorted as one identity cluster. At the following training stage, the bases of the age and identity subspace are learnt from the prepared data, where the dimensions of identity and age factors are set to 10 and 100 respectively. When the dictionary which conveys desired age properties is built up from the obtained age components, we warp the $K_{i}$ atoms in it to the same configuration as the probe face for sparse reconstruction, thus generating aging effects on faces. More details in parameter settings are shown in Tab. 2.

In our experiments, we comprehensively evaluate the proposed aging model in four layers: (I) aging and rejuvenating simulation on face images; (II) evaluating the accuracy of age synthesis and preservation of identity; (III) comparison with ground-truth; and (IV) comparison with state-of-the-art.

\subsection{Exp. I: Simulation of Face Aging and Rejuvenating Effects}

\begin{figure*}[!t]
\centering
\includegraphics[width=1\textwidth]{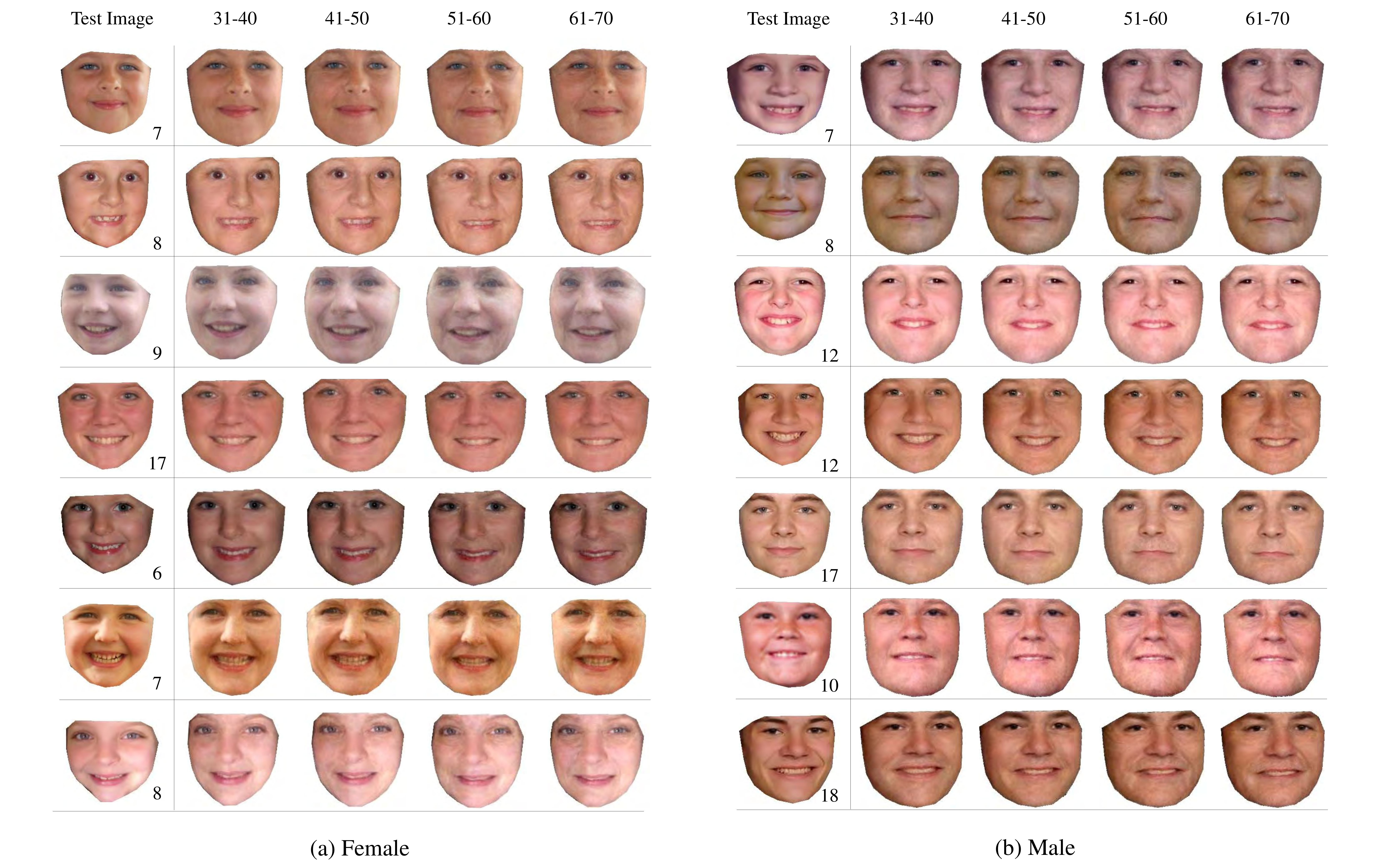}
\caption{Aging simulation results for the subjects in the IRIP database. From left to right, the columns display the probe face and the synthesized results in four sequential age brackets.  (a) Female subjects; (b) Male subjects.}
\label{aging_morph}
\end{figure*}

\begin{figure*}[!t]
\centering
\includegraphics[width=1.02\textwidth]{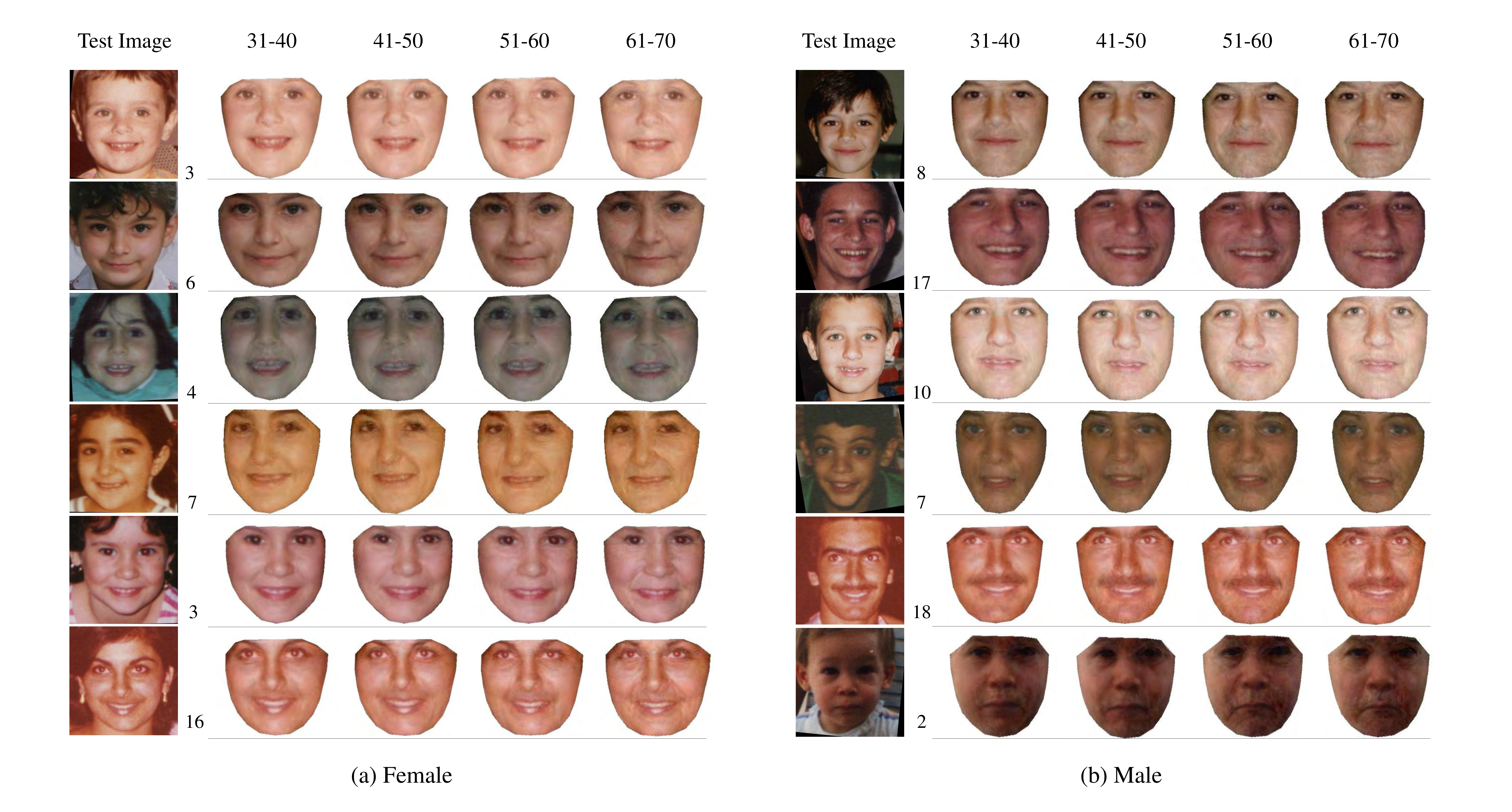}
\caption{Aging simulation results for the subjects in the FG-NET database. From left to right, the columns display the probe face and the synthesized results in four sequential age brackets.  (a) Female subjects; (b) Male subjects.}
\label{aging_fgnet}
\end{figure*}

\begin{figure*}[!t]
\centering
\includegraphics[width=1\textwidth]{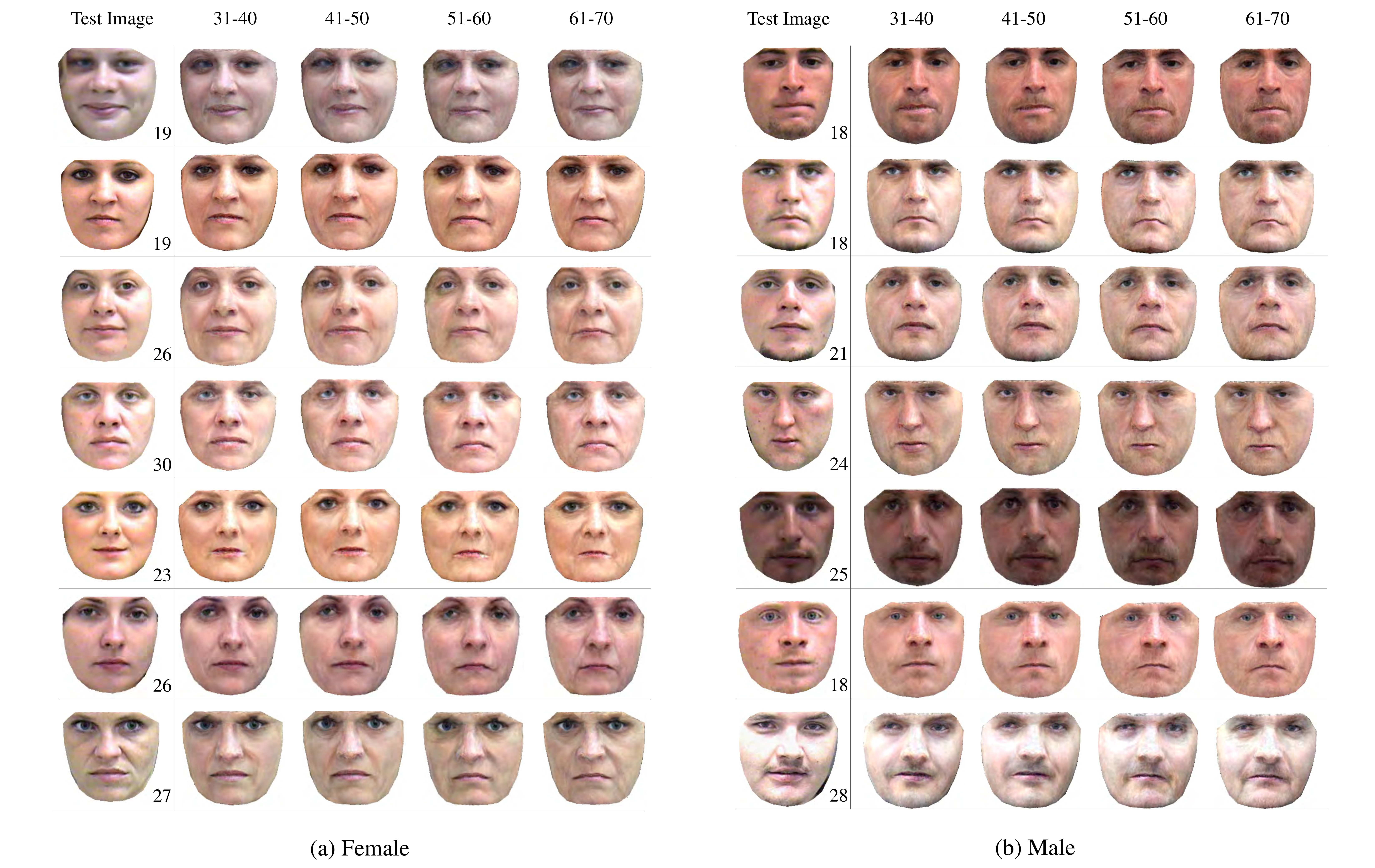}
\caption{Aging simulation results for the subjects in the MORPH database. From left to right, the columns display the probe face and the synthesized results in four sequential age brackets. (a) Female subjects; (b) Male subjects.}
\label{aging_irip}
\end{figure*}

\begin{figure*}[!t]
\setcounter{figure}{8} 
\centering
\includegraphics[width=0.95\textwidth]{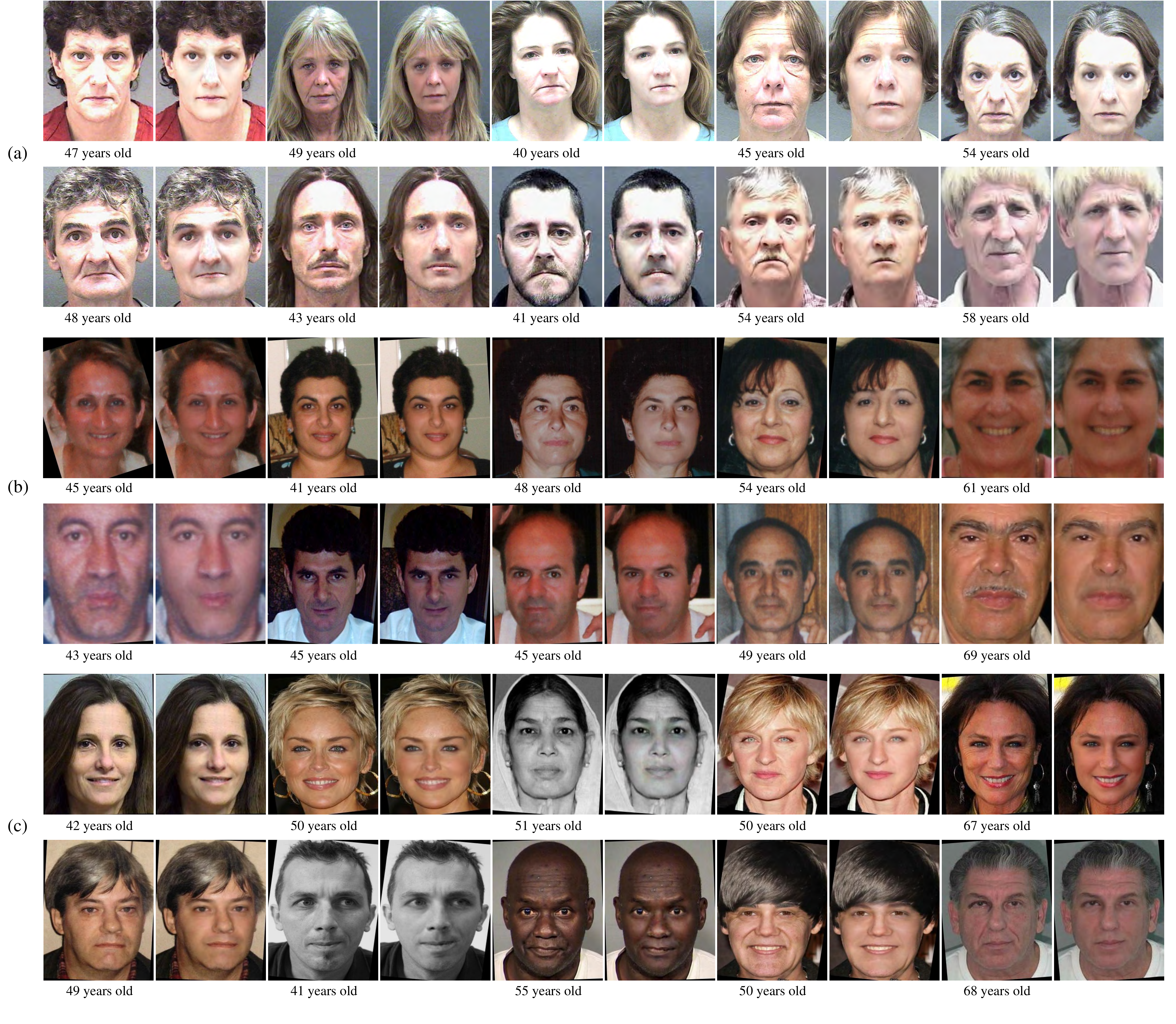}
\caption{Rejuvenating effect synthesis results: (a) The MORPH database; (b) The FG-NET database; (c) The IRIP database.}
\label{Rejuvenating}
\end{figure*}

With the above-mentioned training set, we apply the proposed age progression method on the faces of the IRIP and FG-NET databases ranging from 2 to 18 years old and synthesize a sequence of aging rendering for them, and the simulation results lie in the target age range of [31-40], [41-50], [51-60], and [61-70]. Due to lack of images of teen age, the test faces in MORPH are between 18 and 30 years old. Fig. 2 demonstrates some of the obtained age-processed faces which are blended into the original heads and backgrounds, Fig. 6 shows more results achieved on the IRIP database, and those of FG-NET and MORPH are displayed in Fig. 7 and Fig. 8 respectively. We can see that satisfactory age-specific properties are generated on the faces, regardless of race, skin color and expression, and more winkles emerge as target age grows.

We also carry out rejuvenating experiments to test our algorithm. In this experiment, all of the test faces come from the people older than 40 years old, and they are transformed to the age bracket of [21, 30]. The results of rejuvenating effect simulation are shown in Fig. 9, 9(a), 9(b) and 9(c) exhibit the ones of the MORPH, the FG-NET and the IRIP database respectively.

\subsection{Exp. II: Contributions of Shape Aging Effects }
Facial shape variations by aging can be often observed, especially during early growth. At the child growth stage, the craniofacial deformation is much more drastic compared to facial texture changes, and the face size would increase and forehead slope back. As our model focuses on the long-term age progression, early aging is naturally involved in. Noticing that simulating the growth of profiles significantly influences the rendering results, specially we make a comparison between the synthesized faces with and without shape transformation.

Some of the comparison results are demonstrated in Fig. 5 (b) and 5 (c). We can see from the results that the faces look much more vivid and convincing after shape aging is performed.

\begin{table}[t]
\renewcommand{\arraystretch}{1.3}
\caption{Parameter demonstration in the experiments for aging effects synthesis}
\label{Parameter}
\centering
\begin{tabular}{c|c|c|c|c|c|c}
\hline
\hline
\multicolumn{2 }{c}{\bfseries{Training Image}} & \multicolumn{4 }{|c|}{\bfseries{Hidden Factor Analysis}} & \bfseries{{\tabincell{c}{Sparse\\ Representation}}}\\
\hline
Size (pixel) & $d$ & $p$ & $q$ & $M$ & $N$ & $K_{i}$\\
\hline
100 $\times$ 100& 30000 & 10 & 100 & 7 & 1050 & 150 \\
\hline
\hline
\end{tabular}
\end{table}

\subsection{Exp. III: Aging Model Evaluation}

To quantitatively assess the performance of the proposed aging simulation model, we follow the estimation strategy as adopted in several previous work \cite{5,13,31}, and carry out both subjective and objective experiments to evaluate if the age-specific characteristics are truly added in the synthesized faces as well as whether the identity information of the source face is well retained during age transformation. Additionally, we also compare the age progression results with the ground-truth images and the state of the art.

\subsubsection{Exp. III-A: Evaluating the Ability of Generating Age-Related Characteristics}

Deriving from the emergence of aging effects, the perceived age (the individual age gauged by human subjects from the visual appearance \cite{1}) is supposed to increase in the synthesized face. Correspondingly, we conduct age estimation in this part to assess the accuracy of aging synthesis. Allowing for that the FG-NET database is widely exploited for training automatic age estimation model \cite{20,24,32}, we conduct human observation based evaluation, for averting the problem of overfitting. Specifically, twenty human observers participate in this experiment, and they are asked to label the perceived age of the synthesized faces.

In more detail, 20 young source faces in the FG-NET database are randomly selected, and each of them owns a synthesized aging sequence as Fig. 7 shows; thus 80 corresponding age-progressed faces constitute the testing set in this evaluation, which are supposed to lie in four older age brackets: [31-40], [41-50], [51-60], and [61-70]. The synthesized results of different subjects and target age groups are all mixed together, and we upset the order of the images and present them to the human observers one by one, for a set of precise age value as feedback. The evaluations are separately conducted with regard to the gender attribute of the test samples, and performed on the MORPH and IRIP datasets with the same protocol as on the FG-NET database. 

\begin{figure}[h]
\setcounter{figure}{9} 
\centering
\subfigure[]{
\begin{minipage}[b]{0.225\textwidth}
\includegraphics[height = 1.25\textwidth,width=1\textwidth]{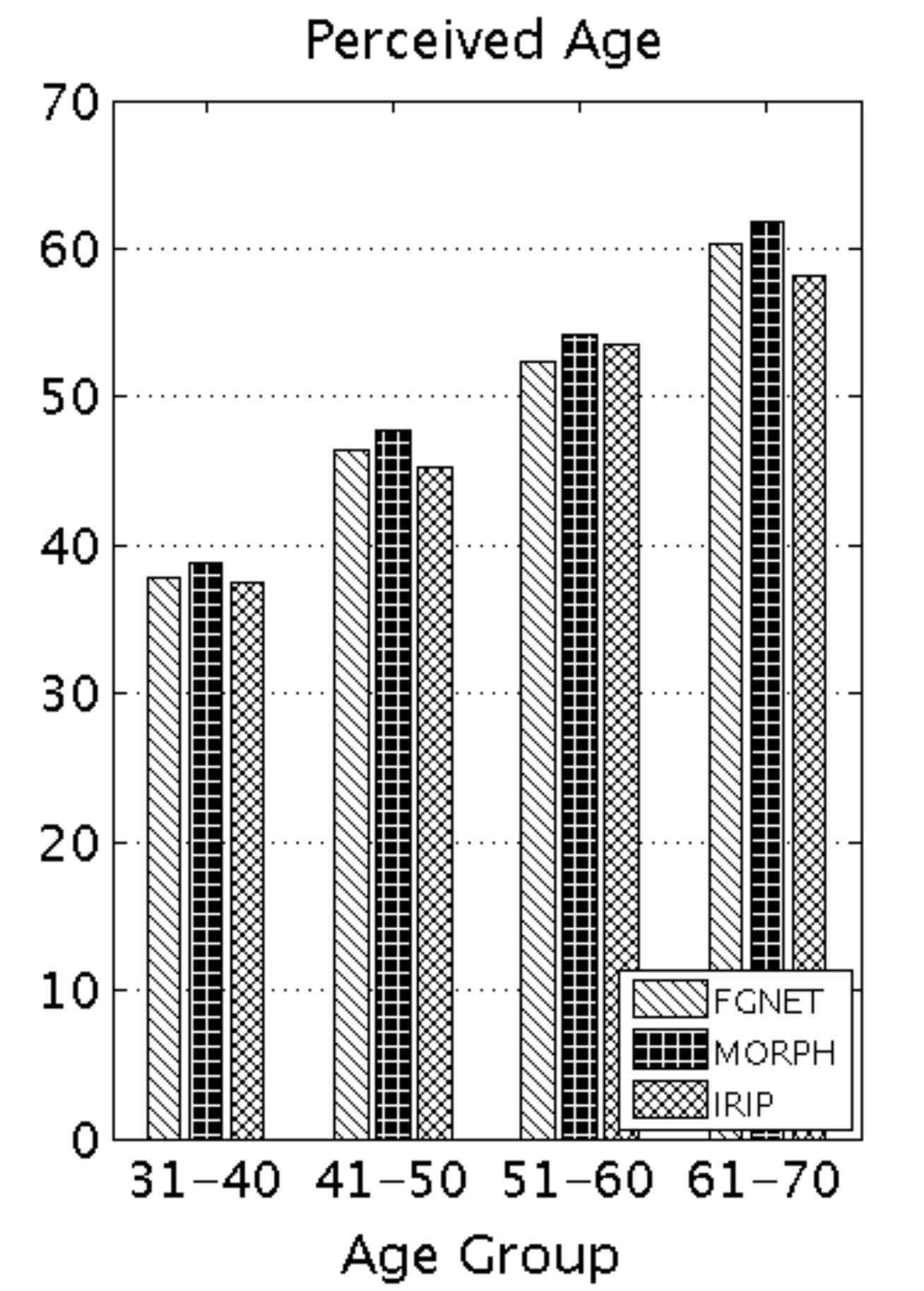} 
\end{minipage}
}
\subfigure[]{
\begin{minipage}[b]{0.225\textwidth}
\includegraphics[height = 1.25\textwidth,width=1\textwidth]{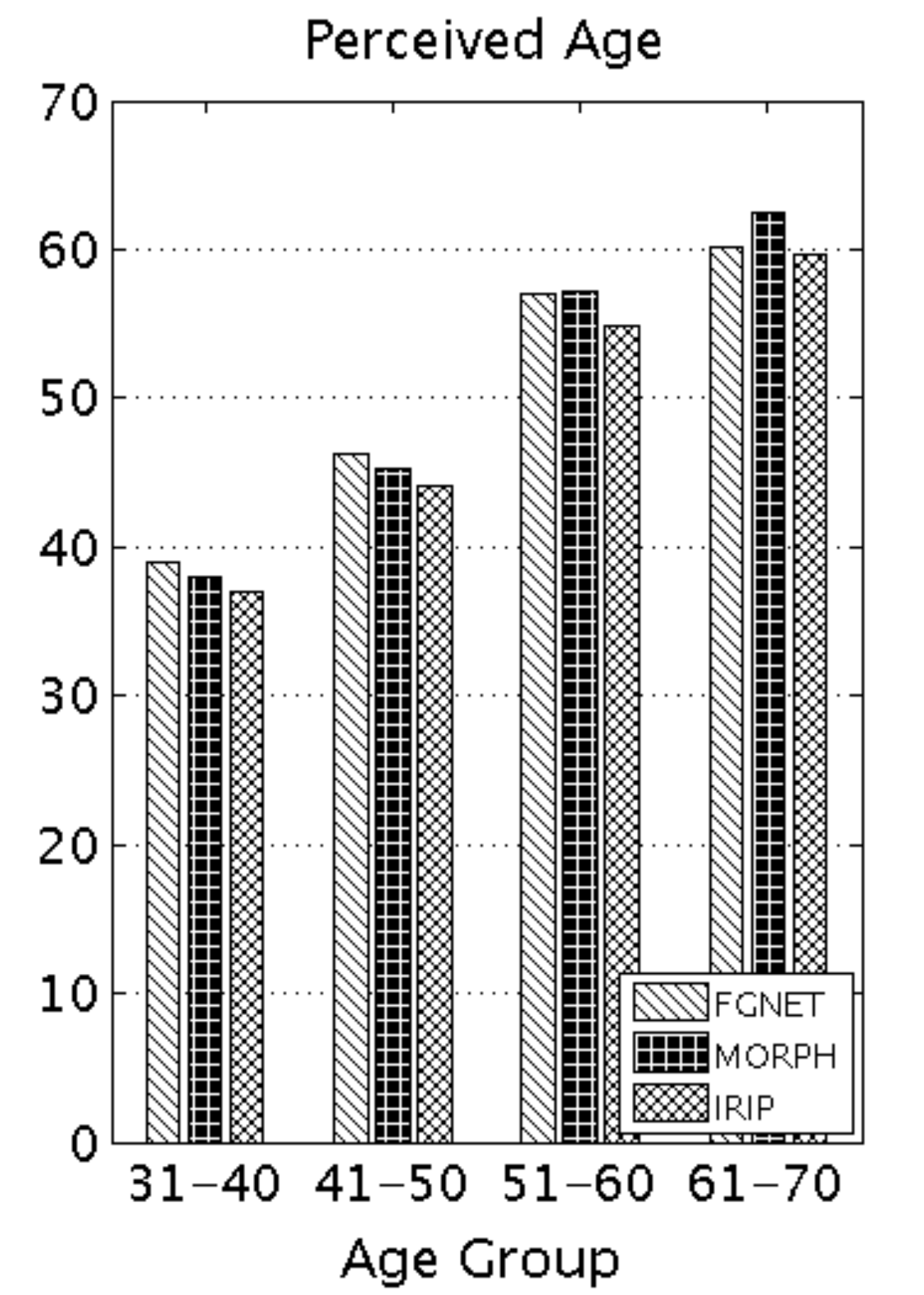} 
\end{minipage}
}
\caption{Human-based age estimation results: (a) Female subjects; (b) Male subjects.}
\end{figure}

Fig. 10 plots the human-based age estimation results achieved on these three datasets. It shows that the perceived age results increase firmly along with the change of age groups, and they conform to the target age in principle, which clearly validate the effectiveness of the proposed aging model. It should be clarified that the averaged estimation results of the target age group [61-70] are relative lower, while that of [31-40] are relative higher, the following could possibly explain this phenomenon: (i) Although the perceived age is defined on the appearance age (the age information shown on the visual appearance \cite{1}), the deviation between them exists objectively; and (ii) Compared with the samples from the two middle age groups whose perceived ages may fluctuate up and down, human observers do not tend to give a score rather low or high to a synthesized image, which probably comes from the marginal age groups.


\subsubsection{Exp. III-B: Evaluating Preservation of Identity}
In this experiment, we attempt to evaluate whether the original identity is well preserved in the age-progressed face image; hence conduct objective face recognition for assessing. 
Similar to the {\em Exp. III-A}, 20 young source faces (under 18 years old) belonging to different subjects are randomly selected from the FG-NET database and adopted as the gallery samples, and we then identify their corresponding synthetic faces of four age groups ({\em i.e.} [31-40], [41-50], [51-60], and [61-70]). To make the evaluation more general, we further add 50 face images to the gallery set as the disturbing samples, which are randomly chosen from another 50 individuals of various age groups. Finally, we use the  face recognition API of Face++ \cite{30} to determine which one of the 70 gallery images is the synthetic image generated from. The evaluations on MORPH and IRIP are performed in similar settings, except that the size of the gallery set increases to 100 (including 50 disturbing images), mainly considering that more face samples are available in these two datasets. The recognition results are shown in Fig. 11.

\begin{figure}[h]
\centering
\includegraphics[width = 3in]{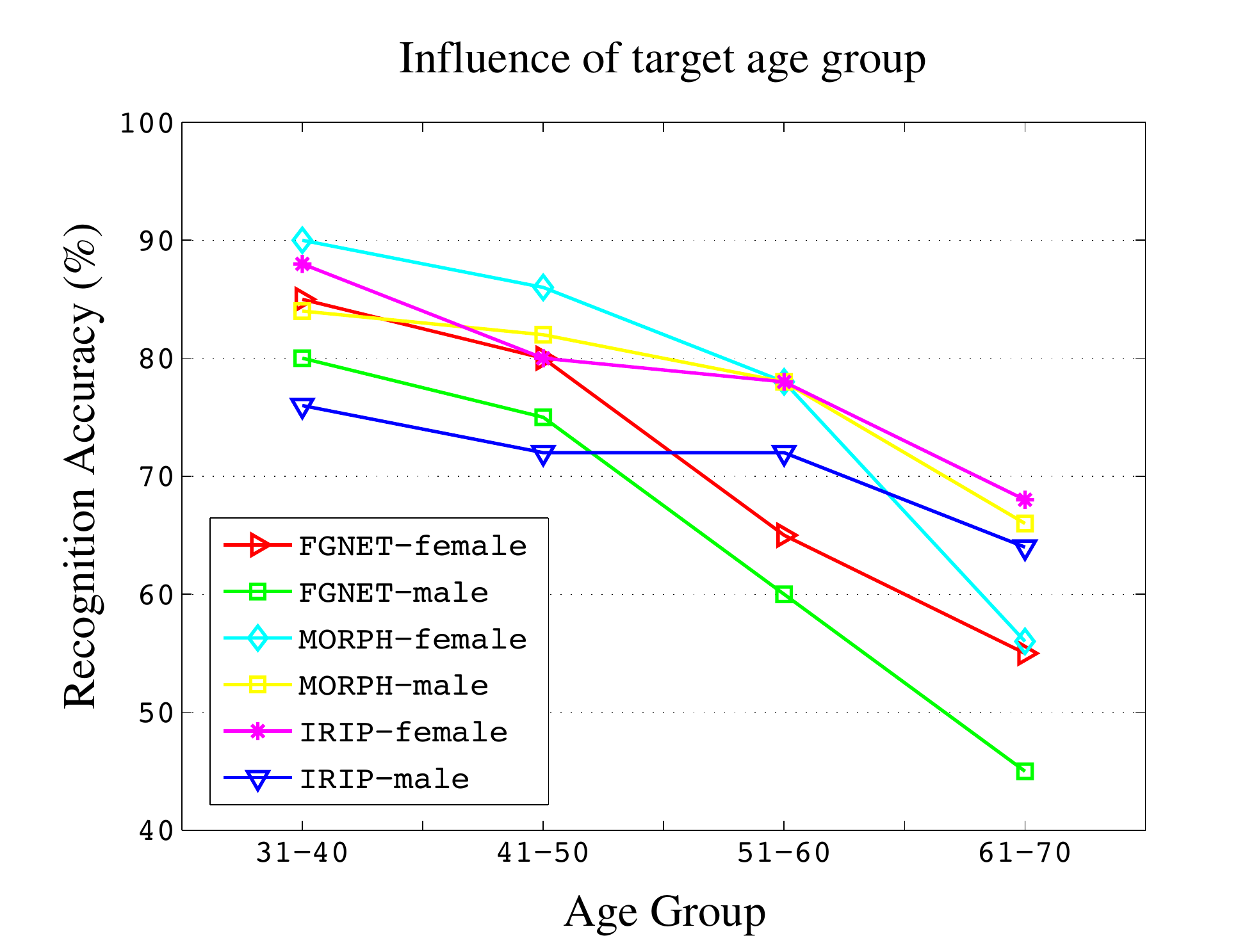}
\caption{Automatic face recognition results based on Face++. }
\label{ComparisonWithGround-truth}
\end{figure}

In the figure, it is obvious that the rank-one recognition rate reaches the maximum value at the target age group of [31-40] for both male and female on the three datasets, and all the results achieved exceed 75\%, verifying the ability of identity preservation of the proposed aging method. In addition, the recognition accuracy decreases as the aging period lengthens, which also conforms to the physical truth of face aging.


\subsubsection{Exp. III-C: Comparison with Ground-truth Images}

\begin{figure}[t]
\centering
\includegraphics[width=3.5in]{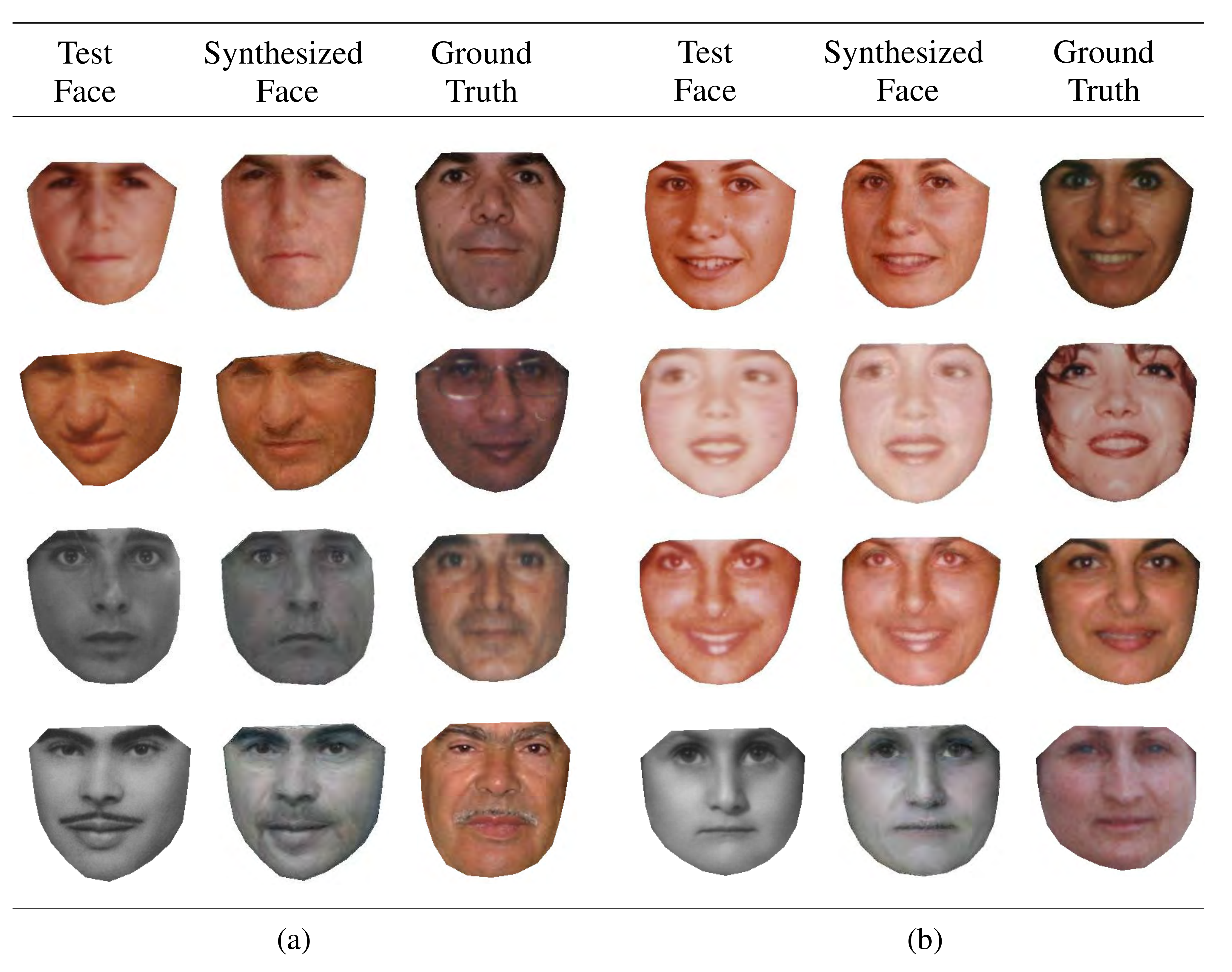}
\caption{Comparison with ground-truth on the FG-NET aging database: (a) Male subjects; (b) Female subjects.}
\label{ComparisonWithGround-truth}
\end{figure}
In order to further estimate the proposed aging model, we perform face recognition on the face images before and after synthesis respectively and compare their accuracies, where the ground-truth images are regarded as the gallery samples. This assessment requires a relative large age gap between the test young face and the ground-truth image of an old face of the same person. However it should be noted that the longitudinal age span of a subject in the MORPH dataset mostly varies from 5 to 10 years, and each subject in IRIP only has a single image. Both the databases are therefore ruled out in this experiment. Only FG-NET is considered here. Among its 82 subjects, 37 of them contain the photos both of the teen age and middle age, and they are used in recognition. For each eligible subject, we have an input young face, the corresponding aging simulation result, and a ground-truth image. Some results are presented in Fig. 12, and in each case from left to right, it displays the test face, the age progression result, and the face image belonging to the same individual at the corresponding age, respectively.

We then employ the commercial software, namely Face++, to conduct face recognition. In one scenario, the 37 ground-truth faces are used as the gallery samples, and the corresponding 37 young faces as the probe samples; while in the contrast scenario, the gallery set is the same and the synthesized aged faces are used as probes. As displayed in Fig. 13, the recognition rate does improve after aging simulation. It illustrates the effectiveness of the proposed method.

\begin{figure}[!t]
\centering
\includegraphics[width=3in]{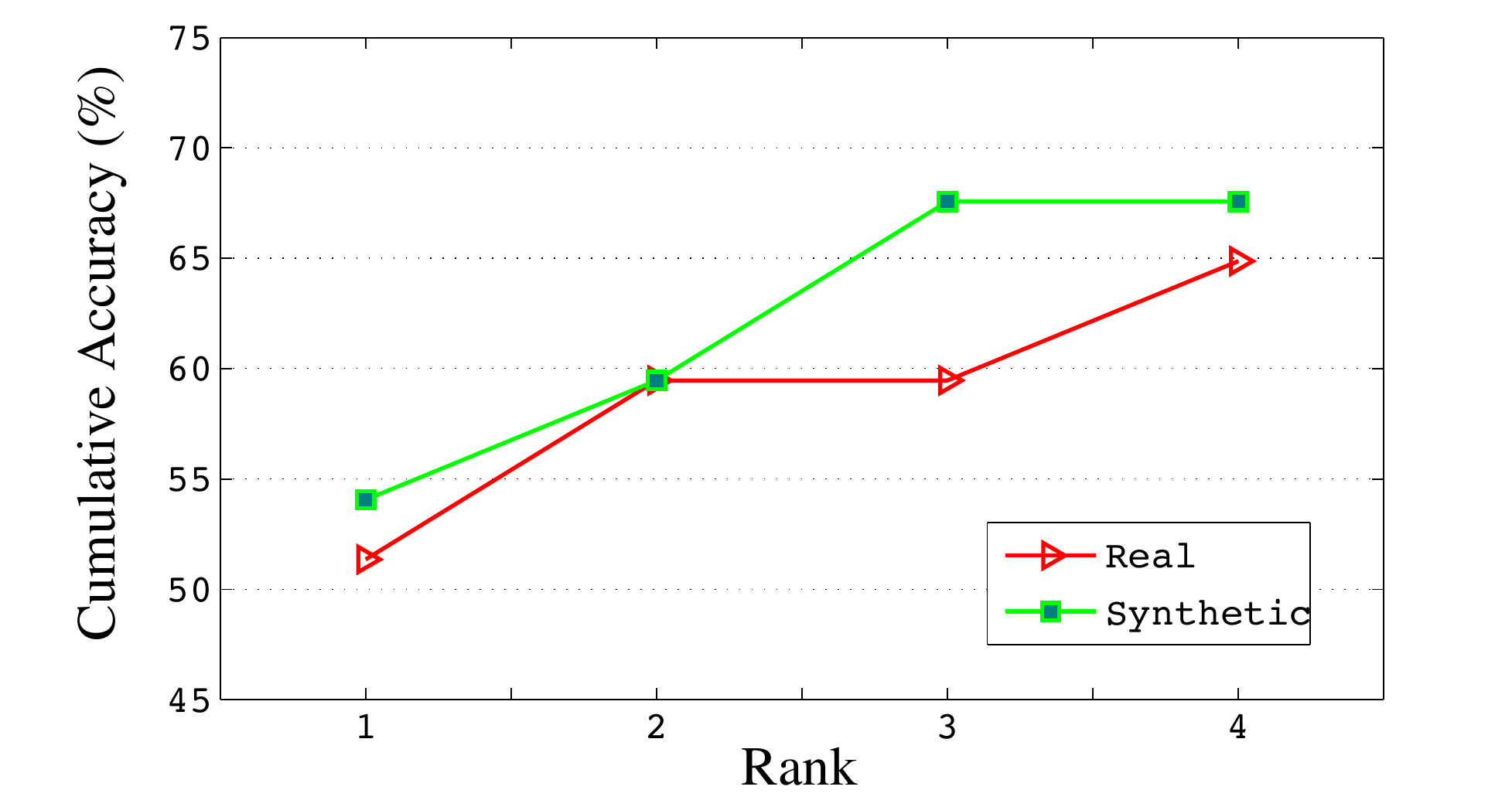}
\caption{Comparison of recognition rates on the face images before and after synthesis.}
\label{ComparisonWithGround-truth2}
\end{figure}

 \begin{table*}[t]
\renewcommand{\arraystretch}{1.3}
\caption{Comparison with prior work in identity preservation and aged facial feature generation}
\label{ComparisonWithPrior}
\centering
\begin{tabular}{c|c|C{1.25cm}|C{1.25cm}|C{1.25cm}|C{1.25cm}|C{1.25cm}|C{1.25cm}}

\hline
\hline
 \multicolumn{2 }{c|}{} &  \multicolumn{2 }{c|}{~~~Park {\em et\ al.} \cite{10}~~~} &  \multicolumn{2 }{c|}{Wang  {\em et\ al.} \cite{13}} &  \multicolumn{2 }{c}{Kemelmacher {\em et\ al.} \cite{9}}  \\ 
\cline{3-8}
\multicolumn{2}{c|}{} &  Age & Identity  &    Age & Identity  &    Age & Identity  \\
\hline
\multirow{2}{*}{{\tabincell{c}{Objective\\ evaluation}}} & Prior work &--& 0.9355 & -- & 0.5148 & -- & 0.8523   \\
\cline{2-8}
& This work &-- & 0.8080 &-- & \bfseries{0.9435}  & -- & \bfseries{0.8696}  \\
\hline
\multirow{2}{*}{{\tabincell{c}{Subjective\\ evaluation}}}& Prior work  &0.4950  & 0.8610 & 0.7975  &  0.6835 & 0.7405 & 0.7854   \\
\cline{2-8}
& This work & \bfseries{0.7995} &  0.7185  & \bfseries{0.8685}  & \bfseries{0.8015}  & \bfseries{0.8186} & 0.7772    \\
\hline
\hline

\end{tabular}
\end{table*}

\begin{figure*}[t]
\centering
\includegraphics[width=0.75\textwidth]{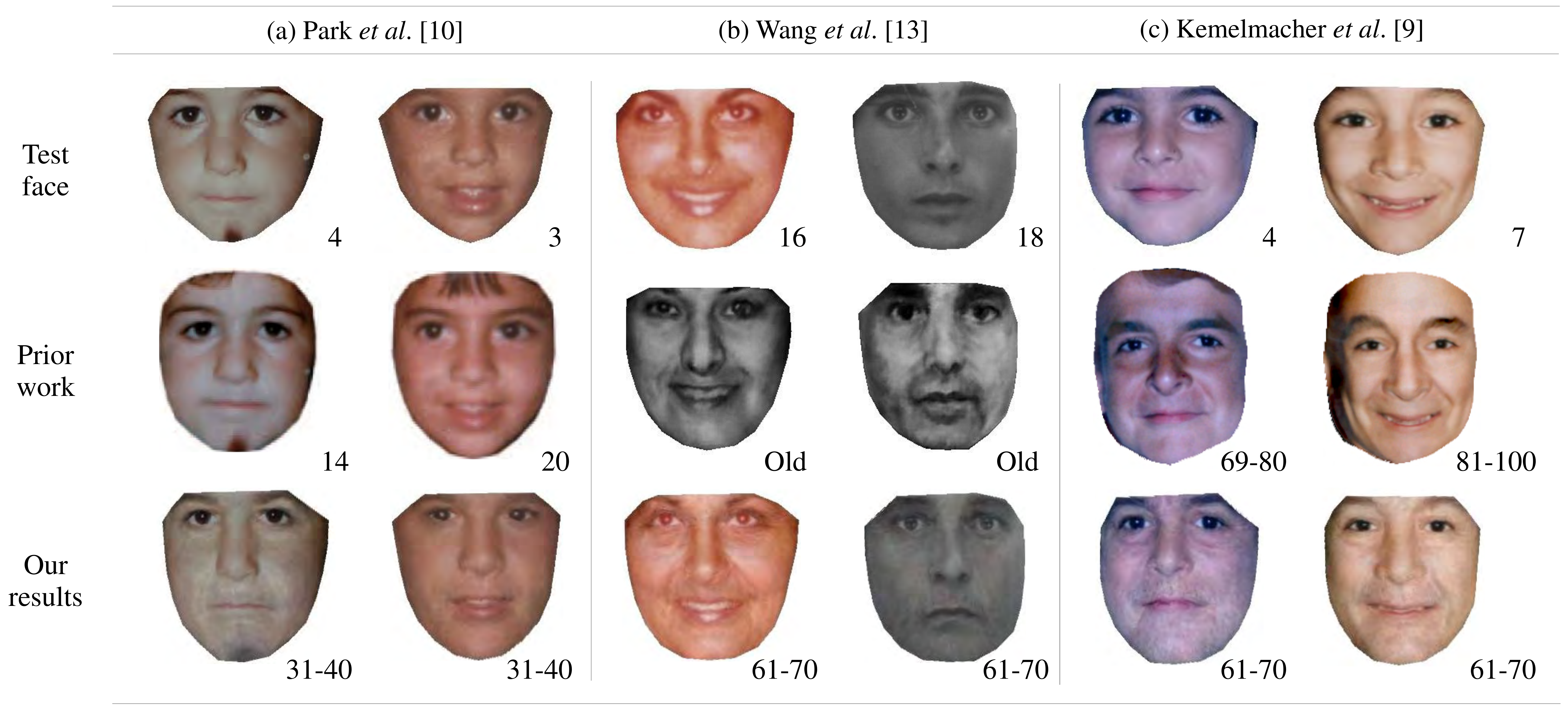}
\caption{Comparison with the prior work on the FG-NET aging database: (a) Park  {\em et\ al.}; (b) Wang  {\em et\ al.}; (c) Kemelmacher  {\em et\ al.}. }
\label{ComparisonWithPrior2}
\end{figure*}

\subsubsection{Exp. III-D: Comparison with State-of-the-art}

In this part, we compare the proposed aging approach with the previous studies in the literature that demonstrate age-progressed faces on the public FG-NET database ({\em i.e.} \cite{9}\cite{10}\cite{13}), which also signify the latest advances in aging synthesis. Different test samples are demonstrated in the previous papers, and we thus select a few used in common, as Fig. 14 shows. The source faces are exhibited in the first row, the synthesized results of the tasks in the literature in the middle row, and the images of the same target ages obtained by the proposed method in the last row.

This comparison is performed based on both objective and subjective evaluations. Unfortunately, in objective evaluation, quantitative comparison cannot be directly achieved by face recognition, since the synthesized faces provided in the three tasks are quite limited. Instead, we quantitatively compare the performance according to the similarity between the face images before and after simulation, to see whether identity information is changed. The API of Face++ is used to calculate the similarity value of two given faces. In subjective evaluation, the human observers are asked to rate the synthesized faces whose scores range from 0 to 1 in two aspects, {\em i.e.} aging effect generation and identity preservation.

In this case, 18 different samples shown in \cite{9}\cite{10}\cite{13} are investigated, and the exact figures are listed in Tab. 3. From this table, we can observe that in objective evaluation, the similarity value reached by our method is higher than that in \cite{13} and \cite{9}, while in subject evaluation, this value of our method is slightly inferior but still comparable to the one in \cite{9} (the difference is less than 1\%), but the scores on aging effect generation are largely superior to those in  \cite{13} and \cite{9} (around 7.1\% and 7.8\% higher than \cite{13} and \cite{9} respectively). We also notice that the score of identity similarity of our method keeps a certain distance from that in \cite{10} in both objective and subjective evaluations (12.7\% and 14.3\% lower respectively). However, we can find out that such a gap is due to the fact that the simulated faces in \cite{10} lack of aged texture and are a little bit far from the target age claimed, which is also confirmed by the big difference (about 30.5\%) between the values of generating aging effects in subjective evaluation. To sum up, we can say that the proposed method outperforms the three counterparts.

\section{Discussion}
Compared with the approaches to aging simulation in the literature, this study shows a very different but effective solution to the core issue, {\em i.e.} face synthesis with aged texture.
Most of the recently proposed age progression approaches are data-based where aging patterns and age-related features are learnt from the training samples, since they show obvious advantages over the others. The learnt aged feature is not a fixed mask nor a single average and it possesses good diversity and provides a large number of aging pattern options for facial aging progress. Our method is data based, and in contrast to the others in this category, we propose to decompose the age information ({\em e.g.} the skin folds and wrinkles) from intrinsic person dependent cues, which helps to keep the identity preserved in the synthesized face. Additionally, this operation is also critical to rejuvenating, which particularly requires that no superfluous texture information is left on the generated young face. 

At the stage of sparse reconstruction based texture synthesis, our method selects the most suitable age components of the target age group for each input test face. These age components well complement each other and their linear combination generates expected aging effects. Meanwhile, the proposed method also adapts to various test samples in skin color, lighting condition, {\em etc}. Thanks to the large number of training samples that provide sufficient and diverse age components, the aging effects achieved are natural and conspicuous. By contrast, the existing aging function based approaches cannot make accurate definition of skin fold growth, and the aging results are not as good as ours.

Even though, there still remain some issues to solve. For some test samples that possess severe pose variations or the light shines onto one side of the face, the generated wrinkles might look relatively messy and the aged faces look not so decent. Besides, compared with the complex biological face aging process of human beings, the test image in face aging simulation is generally of an unknown person (not appear in the training set), and it is also impossible to obtain the information of external factors, such as health condition, living style, working environment and sociality, from the single input image, both of which determine that the aging model cannot be personalized. Therefore, almost all the current studies simplify the problem which only considers the internal factors, {\em i.e.} gender, ethnicity, {\em etc}. The proposed method tries to provide more reliable results in terms of identity preservation as well as aging effect generation by decomposing the identity and age component into individual subspaces. The external factors are also incorporated into the age part for simplicity. More sophisticated aging model might depend on an elaborated training set of sufficient and specific samples, especially on the external factors, which is also a major direction in our future work.

\section{Conclusion}
In this paper, we present a novel approach to face aging effect simulation using Hidden Factor Analysis joint sparse representation. The proposed aging method attempts to separate out the age-specific facial clues that change gradually over time from the stable person-dependent properties, and then merely operates on the age component, by transforming it to a target age group via sparse reconstruction, to present aging effects on faces. The aging and rejuvenating results achieved on three databases indicate that the generated aging rendering, including both the texture and shape changes, are natural and convincing, even if the probe face undergoes variations in expression and skin color. Moreover, a series of evaluations prove that our aging model conform to another essential criterion in face age progression, {\em i.e.} the identity information of the probe face is well-preserved.


\ifCLASSOPTIONcaptionsoff
  \newpage
\fi


\begin{thebibliography}{99}

\bibitem{1}
Y.~Fu, G.~Guo, and T.S.~Huang, ``Age Synthesis and Estimation via Faces: A Survey," \textit{IEEE Transactions on Pattern Analysis and Machine Intelligence}, vol. 32, no. 11, pp. 1955-1976, Nov. 2010.
\bibitem{2}
N.~Ramanathana, R.~Chellappa, and S.~Biswas, ``Computational Methods for Modeling Facial Aging: A Survey," \textit{J. Visual Languages and Computing}, vol. 20, no. 3, pp. 131-144, 2009.
\bibitem{3}
B. P.~Tiddeman, M. R.~Stirrat, and D. I. Perrett, ``Towards Realism in Facial Image Transformation: Results of a Wavelet MRF Method," \textit{Computer Graphics Forum}, vol. 24, no. 3, pp. 449-456, 2005.
\bibitem{4}
A.~Lanitis, C. J.~Taylor, and T. F.~Cootes, ``Toward automatic simulation of aging effects on face images," \textit{IEEE Transactions on Pattern Analysis and Machine Intelligence}, vol. 24, no. 4, pp. 442-455, Apr. 2002.
\bibitem{5}
J.~Suo, S.~Zhu, S.~Shan, and X.~Chen, ``A compositional and dynamic model for face aging," \textit{IEEE Transactions on Pattern Analysis and Machine Intelligence}, vol. 32, no. 3, pp. 285-401, March. 2010.
\bibitem{6}
J.~Suo, F.~Min, S.~Zhu, S.~Shan, and X.~Chen, ``A Multi-Resolution Dynamic Model for Face Aging Simulation," \textit{Proc. IEEE Int'l Conf. Computer Vision and Pattern Recognition}, 2007, pp. 1-8.
\bibitem{7}
N.~Ramanathan and R.~Chellappa, ``Modeling age progression in young faces," \textit{Proc. IEEE Int'l Conf. Computer Vision and Pattern Recognition}, Jun. 2006, pp. 387-394.
\bibitem{8}
T.F.~Cootes, G.J.~Edwards, and C.J.~Taylor, ``Active Appearance Models," \textit{IEEE Transactions on Pattern Analysis and Machine Intelligence}, vol. 23, no. 6, pp. 681-685, June 2001.
\bibitem{9}
I.~Kemelmacher-Shlizerman, S.~Suwajanakorn, and S.M.~Seitz, ``Illumination-aware Age Progression," \textit{Proc. IEEE Int'l Conf. Computer Vision and Pattern Recognition}, pp. 3334-3341, 2014.
\bibitem{10}
U.~Park, Y.~Tong, and A. K. Jain, ``Age-invariant face recognition," \textit{IEEE Transactions on Pattern Analysis and Machine Intelligence},  vol. 32, no. 5, pp. 947-954, May 2010.
\bibitem{11}
U.~Park, Y.~Tong, and A. K. Jain, ``Face recognition with temporal invariance: A 3D aging model," \textit{IEEE International Conference on Automatic Face and Gesture Recognition (FG)}, 2008, pp. 1-7.
\bibitem{12}
F.~Jiang and Y.~Wang, ``Facial aging simulation based on super-resolution in tensor space," \textit{Proc. 15th Int'l Conf. Image Process.}, Oct. 2008, pp. 1648-1651.
\bibitem{13}
Y.~Wang, Z.~Zhang, W.~Li, and F.~Jiang, ``Combining Tensor Space Analysis and Active Appearance Models for Aging Effect Simulation on Face Images," \textit{IEEE transactions on systems, man, and cybernetics. Part B, Cybernetics}, vol. 42, no. 4, pp. 1107-1118, August 2012.
\bibitem{14}
Y. H.~Kwon and N. D.V.~Lobo, ``Age classification from facial images," \textit{Computer Vision and Image Understanding}, vol. 74, no. 1, pp. 1-21, April 1999.
\bibitem{15}
N.~Ramanathan and R.~Chellappa, ``Modeling shape and textural variations in aging faces," \textit{IEEE International Conference on Automatic Face \& Gesture Recognition}, 2008, pp. 1-8.
\bibitem{16}
J.T.~Todd, L.S.~Mark, R.E.~Shaw, and J.B.~Pittenger, ``The perception of human growth," \textit{Scientific American}, vol. 242, no. 2, pp. 132-144, 1980.
\bibitem{17}
H.~Yang, D.~Huang, and Y.~Wang,  ``Age Invariant Face Recognition based on Texture Embedded Discriminative Graph Model,"   \textit{IEEE/IAPR International Joint Conference on Biometrics (IJCB)}, Florida, USA, 2014.
\bibitem{18}
D.~Gong, Z.~Li, D.~Lin, J.~Liu, and X.~Tang, ``Hidden Factor Analysis for Age Invariant Face Recognition," \textit{IEEE International Conference on Computer Vision (ICCV)}, 2013, pp. 2872-2879.
\bibitem{19}
M.~Yang, L.~Zhang, J.~Yang, and D.~Zhang, ``Metaface learning for sparse representation based face recognition," \textit{17th IEEE International Conference on Image Processing (ICIP)}, 2010, pp. 1601-1604.
\bibitem{20}
X.~Geng, Z.~Zhou, K.~Smith-Miles, ``Automatic Age Estimation Based on Facial Aging Patterns,'' \textit{IEEE Transactions on Pattern Analysis and Machine Intelligence}, vol. 29, no. 12, pp. 2234-2240, 2007.
\bibitem{21}
J.~Wright, Y.~Ma, J.~Mairal, S.~Guillermo, T.~Huang, and S.~Yan, ``Sparse Representation for Computer Vision and Pattern Recognition," \textit{Proceedings of the IEEE}, vol. 98, no. 6, pp. 1031-1044, 2010.
\bibitem{22}
J.~Wright, A.~Yang, A.~Ganesh, S.~Sastry, and Y.~Ma, ``Robust Face Recognition via Sparse Representation," \textit{IEEE Transactions on Pattern Analysis and Machine Intelligence}, vol. 31, no. 2, pp. 210-227, Feb. 2009.
\bibitem{23}
M.~Aharon, M.~Elad, and A. M.~Bruckstein, ``K-SVD: An algorithm for designing overcomplete dictionaries for sparse representations," \textit{IEEE Transactions on Signal Processing}, vol. 54, no. 11, pp. 4311-4322, Nov. 2006.
\bibitem{24}
G.~Guo, Y.~Fu, T.S.~Huang, and C.~Dyer, ``A Probabilistic Fusion Approach to Human Age Prediction,'' \textit{Proc. IEEE Computer Vision and Pattern Recognition-Semantic Learning and Applications in Multimedia Workshop}, 2008.
\bibitem{25}
A.J.~O'Toole, T.~Price, T.~Vetter, J.C.~Bartlett, and V.~Blanz, ``3D Shape and 2D Surface Textures of Human Faces: The Role of `Averages' in Attractiveness and Age," \textit{Image and Vision Computing}, vol. 18, pp. 9-19, 1999.
\bibitem{26}
J.~Yang, J.~Wright, T.~Huang, and Y.~Ma, ``Image super-resolution as sparse representation of raw image patches," \textit{IEEE Conference on Computer Vision and Pattern Recognition (CVPR)}, 2008, pp. 1-8.
\bibitem{27}
M.~Protter and M.~Elad. ``Image Sequence Denoising via Sparse and Redundant Representations,'' \textit{IEEE Transactions on Image Processing}, vol. 18, no. 1, pp. 27-35, 2009.
\bibitem{28}
O.~Bryt and M.~Elad, ``Compression of Facial Images Using the K-SVD Algorithm,'' \textit{Journal of Visual Communication and Image Representation}, vol. 19, no. 4, pp. 270-283, 2008.
\bibitem{29}
A.M.~Albert, K.~Ricanek Jr., and E. Patterson, ``A Review of the Literature on the Aging Adult Skull and Face: Implications for Forensic Science Research and Applications," \textit{J. Forensic Science International}, vol. 172, no. 1, pp. 1-9, Apr. 2007.
\bibitem{30}
Megvii Inc. Face++ Research Toolkit. www.faceplusplus.com, December 2013.
\bibitem{31}
A.~Lanitis, ``Evaluating the performance of face-aging algorithms,"  in \textit{Proc. 8th Int. Conf. Autom. Face Gesture Recog.}, 2008, pp. 1-6.
\bibitem{32}
G.~Guo, Y.~Fu, C.~Dyer, and T.S.~Huang, ``Image-Based Human Age Estimation by Manifold Learning and Locally Adjusted Robust Regression,'' \textit{IEEE Trans. Image Processing}, vol. 17, no. 7, pp. 1178-1188, July 2008.
\bibitem{33}
J.~Suo, X.~ChenS, S.~Shan, W.~Gao and Q.~Dai, ``A Concatenational Graph Evolution Aging Model," \textit{IEEE Transactions on Pattern Analysis and Machine Intelligence}, vol. 34, no. 11, pp. 2083-2096, Nov. 2012.
\bibitem{34}
Y.~Shan, Z.~Liu, and Z.~Zhang, ``Image-Based Surface Detail Transfer," \textit{Proc. IEEE Conf. Computer Vision and Pattern Recognition}, pp. 794-799, 2001.
\bibitem{35}
B.P.~Tiddeman, D.M.~Burt, and D.I.~Perrett, ``Prototyping and Transforming Facial Textures for Perception Research," \textit{IEEE Computer Graphics and Applications}, vol. 21, no. 5, pp. 42-50, Sept./ Oct. 2001.
\bibitem{36}
L. S.~Mark, J. T.~Todd, and R. E.~Shaw, ``Perception of growth: A geometric analysis of how different styles of change are distinguished," \textit{J. Exp. Psychol.: Human Percept. Perform.}, vol. 7, no. 4, pp. 855-868, Aug. 1981.
\bibitem{37}
Y.~Wu, N.M.~Thalmann, and D.~Thalmann, ``A Plastic-Visco-Elastic Model for Wrinkles in Facial Animation and Skin Aging," \textit{Proc. Second Pacific Conf. Fundamentals of Computer Graphics}, pp. 201-213, 1994.
\bibitem{38}
Y.~Wu, N.M.~Thalmann, and D.~Thalmann, ``A Dynamic Wrinkle Model in Facial Animation and Skin Aging," \textit{J. Visualization and Computer Animation}, vol. 6, no. 4, pp. 195-205, 1995.
\bibitem{39}
The FG-NET aging database. http://www.fgnet.rsunit.com/.
\bibitem{40}
 K.~Ricanek Jr, and T.~Tesafaye, ``Morph: A Longitudinal Image Database of Normal Adult Age-Progression," \textit{Proc. Seventh International Conf. Automatic Face and Gesture Recognition}, pp. 341-345, 2006.
\end{thebibliography}
\end{document}